\documentclass[10pt,twocolumn,letterpaper]{article}
\pdfoutput=1

\usepackage{cvpr}
\usepackage{times}
\usepackage{epsfig}
\usepackage{graphicx}
\usepackage{amsmath}
\usepackage{amssymb}
\usepackage{multirow}
\usepackage{booktabs} 
\usepackage{mathtools}
\usepackage[normalem]{ulem}
\usepackage{tabu}
\usepackage{subfigure}
\usepackage[ruled,linesnumbered]{algorithm2e}
\usepackage{xfrac}
\usepackage{caption}
\usepackage{color}
\usepackage{dblfloatfix}
\usepackage{float}

\newcommand\Set[2]{\{\,#1\mid#2\,\}}

\definecolor{archi_loss}{RGB}{228,108,10}
\definecolor{archi_e}{RGB}{0,32,96}
\definecolor{archi_dd}{RGB}{225,162,0}
\definecolor{archi_dn}{RGB}{0,176,240}
\definecolor{archi_dl}{RGB}{187,202,40}
\definecolor{archi_dm}{RGB}{107,107,107}
\definecolor{archi_sa}{RGB}{149,55,53}

\newcommand{\beginsupplement}{%
        \setcounter{table}{0}
        \renewcommand{\thetable}{S\arabic{table}}%
        \setcounter{figure}{0}
        \renewcommand{\thefigure}{S\arabic{figure}}%
     }

\makeatletter
\newcommand\refwithdefault[2]{%
  \@ifundefined{r@#1}{%
    #2%
  }{%
    \ref{#1}%
  }%
}
\makeatother

\newcommand{\comment}[1]{}

\definecolor{orange}{RGB}{255,127,0}

\usepackage[pagebackref=true,breaklinks=true,letterpaper=true,colorlinks,bookmarks=false]{hyperref}

\cvprfinalcopy 

\def\cvprPaperID{305} 

\ifcvprfinal\fi

\graphicspath {{./figures/}}
\begin{document}


\title{Seeing Beyond Appearance -- Mapping Real Images into Geometrical Domains\\for Unsupervised CAD-based Recognition} 

\makeatletter
\renewcommand*{\@fnsymbol}[1]{\ensuremath{\ifcase#1\or $CO$\or \dagger\or \ddagger\or
		\mathsection\or \mathparagraph\or \|\or **\or \dagger\dagger
		\or \ddagger\ddagger \else\@ctrerr\fi}}
\makeatother
\author{
	Benjamin Planche\thanks{These authors contributed equally to the work.}\hspace{1em}$^{,1,2}$, Sergey Zakharov\footnotemark[1]\hspace{1em}$^{,1,3}$, 
	Ziyan Wu$^{1}$, \\Andreas Hutter$^{1}$, Harald Kosch$^{2}$, Slobodan Ilic$^{1,3}$\\
	$^{1}$Siemens Corporate Technology\\
	{\tt\small \{benjamin.planche, andreas.hutter, slobodan.ilic, ziyan.wu\}@siemens.com}
	\and
	$^{2}$University of Passau\\
	{\tt\small harald.kosch@uni-passau.de}
	\and
	$^{3}$Technical University of Munich\\
	{\tt\small sergey.zakharov@tum.de}
}

\maketitle

\begin{abstract}
While convolutional neural networks are dominating the field of computer vision, one usually does not have access to the large amount of domain-relevant data needed for their training.
It thus became common to use available synthetic samples along domain adaptation schemes to prepare algorithms for the target domain.
Tackling this problem from a different angle, we introduce a pipeline to map unseen target samples into the synthetic domain used to train task-specific methods. 
Denoising the data and retaining only the features these recognition algorithms are familiar with, our solution greatly improves their performance.
As this mapping is easier to learn than the opposite one (\ie to learn to generate realistic features to augment the source samples), we demonstrate how our whole solution can be trained purely on augmented synthetic data, and still perform better than methods trained with domain-relevant information (\eg real images or realistic textures for the 3D models). Applying our approach to object recognition from texture-less CAD data, we present a custom generative network which fully utilizes the purely geometrical information to learn robust features and achieve a more refined mapping for unseen color images.
\end{abstract}

\begin{figure*}[h!]
  \centering
  \includegraphics[width=\linewidth]{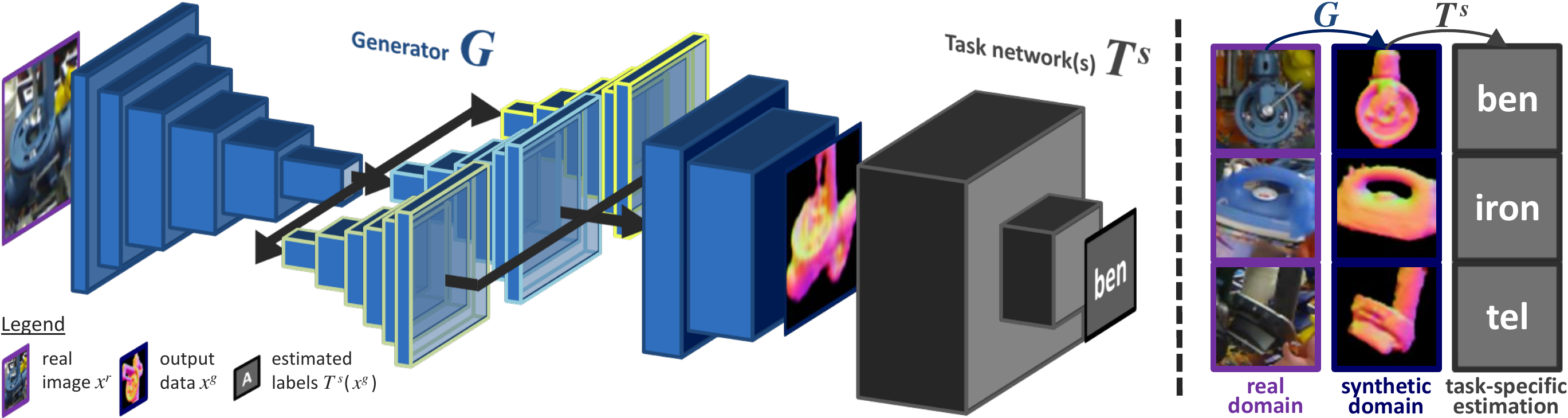}
  \caption{\textbf{Pipeline Usage}. We present $G$, a custom generative network which maps real unseen data into discriminative synthetic domains $X^s$ available for training (\eg normal maps rendered from provided CAD models).
  The pre-processed data can then be handed to any recognition methods $T^s$, themselves simply trained on $X^s$, to achieve high performance despite the lack of any domain-relevant training information.}
  \label{fig:testing}  
\end{figure*}

\section{Introduction}

The ever-increasing popularity of deep convolutional neural networks seems well-deserved, as they are adopted for more and more complex applications.
This success has to be slightly nuanced though, as these methods usually rely on large annotated datasets for their training.
In many cases still (\eg for scalable industrial applications), it would be extremely costly, if not impossible, to gather the required data.
For such use-cases and many others, synthetic models representing the target elements are however usually pre-available (industrial 3D CAD blueprints, simulation models, \etc). It thus became common to leverage such data to train recognition methods \eg by rendering huge datasets of relevant synthetic images and their annotations.

However, the development of exhaustive, precise models behaving like their real counterparts is often as costly as gathering annotated data (\eg acquiring precise texture information, to render proper images from CAD data, actually imply capturing and processing images of target objects). As a result, the salient discrepancies between model-based samples and target real ones (known as \textit{realism gap}) still heavily impairs the application of synthetically-trained algorithms to real data. Research in \textit{domain adaptation} thus gained impetus the last years. Several solutions have been proposed, but most of them require access to real relevant data (even if unlabeled) or access to synthetic models too precise for scalable real-world use-cases (\eg access to realistic textures for 3D models).

In our work, we introduce a novel approach, \textit{SynDA}, tackling domain adaptation and realism gap from a different angle. \textit{SynDA} is composed of a custom generative network to map unseen real samples toward a relevant, easily-available synthetic domain (\eg normal maps), in order to improve recognition for methods themselves trained on this noiseless synthetic modality. 
Along this paper, we demonstrate that it is sensible to train task-specific networks on noiseless information so they learn clean discriminative features, and then develop a mapping function from real to synthetic data; rather than to focus on developing or learning pseudo-realistic noise models to train against (though the two can be complementary to bridge the gap both ways). 

Applied in this paper to CAD-based recognition in color pictures, our approach is based on the assumptions that real-world images can be mapped to the synthetic domain; and that, in absence of any real training data, this mapping can be learned by recovering the synthetic samples altered by a stochastic noise source.
Since our method only needs to eliminate noise and retain features, it performs better than usual generative solutions for domain adaptation, which learns the more difficult task of generating complex features to mimic the target data.
As long as the synthetic domains contain all relevant features and as long as those features are contained in real images, our approach successful enhances recognition, as demonstrated through our empirical evaluation. In summary, we are making the following contributions:

\par\noindent
\textbf{(a) Synthetic modality regression for domain adaptation} --  We propose a novel framework to learn a mapping from unseen real data to relevant synthetic domains, denoising and recovering the information needed for further recognition. Our solution thus not only covers the real-synthetic gap, but also takes care of cross-modality mapping. More specifically, we present how color images can be mapped to normal maps, to help pose-regression and classification in absence of reliable texture information for training.

\par\noindent
\textbf{(b) Decoupling domain adaptation from recognition} -- Most domain adaptation schemes constrain the methods training, by adding pseudo-realistic or noisy features to their training set, editing their architecture or losses, \etc. In our framework, task-specific algorithms simply learn on available, relevant synthetic data (clean normal maps from CAD models in our case), while separately our network $G$ is trained on noisy data to map them into the selected synthetic domain. This decoupling makes training more straightforward, and allows $G$ to be used along any number of recognition methods.
We furthermore observe better results compared to recognition methods directly trained on augmented data. Results even compare to solutions using real information for training.

\par\noindent
\textbf{(c) Performance in complete absence of real training data} -- 
Domain adaptation approaches usually assume the realism gap to be already partially bridged, requiring access to some target domain images or realistic synthetic models (\eg textured 3D data). Opting for recognition tasks with texture-less CAD data for only prior, we demonstrate how our pipeline can be trained on purely synthetic data and still generalize well to real situations. For that we leverage an extensive augmentation pipeline, used as a noise source applied to training samples so our solution learns to denoise and retain the relevant features.

\par\noindent
\textbf{(d) Multi-task metwork with self-attentive distillation} -- The one advantage of synthetic models such as CAD data is the possibility to easily extract various precise modalities and ground-truths to learn from. We thus consolidate several state-of-the-art works~\cite{kuga2017multi,kendall2017multi,xu2018pad,zhang2018self} to develop a custom generator with multiple convolutional decoders for each relevant synthetic modality (\eg normal maps, semantic masks, \etc), and a distillation module on top making use of self-attention maps to refine the final outputs.

After providing a pertinent survey in Section~\ref{sec:rw} and describing our methodology in Section~\ref{sec:mth}, we evaluate our solution in Section~\ref{sec:exp} over a set of different recognition methods and datasets to support our claims.

\section{Related Work}
\label{sec:rw}

Domain adaptation became an increasingly present challenge with the rise of deep-learning methods. We thus dedicate most of our literature review to listing main solutions developed to bridge the gap between real and synthetic data. In a second time, we present convolutional neural network (CNN) methods for shape regression, as we put emphasis on the mapping from real color images to synthetic geometrical domains in this paper.

\par\noindent
\textbf{Bridging the realism gap:} The realism gap is a very well known problem for computer vision methods that rely on synthetic data, as the knowledge acquired on these modalities usually poorly translates to the more complex real domain, resulting in a dramatic accuracy drop. Several ways to tackle this issue have been investigated so far.
A first obvious solution is to improve the quality and realism of the synthetic models. Several works tries to push forward simulation tools for sensing devices and environmental phenomena. State-of-the-art depth sensor simulators work fairly well for instance, as the mechanisms impairing depth scans have been well studied and can be rather well reproduced 
\cite{landau2015simulating,planche2017depthsynth}. 
In case of color data however, the problem lies not in the sensor simulation but in the actual complexity and variability of the color domain (\eg sensibility to lighting conditions, texture changes with wear-and-tear, \etc). This makes it extremely arduous to come up with a satisfactory mapping, unless precise, exhaustive synthetic models are provided (\eg by capturing realistic textures). Proper modeling of target classes is however often not enough, as recognition methods would also need information on their environment (background, occlusions, \etc) to be applied to real-life scenarios.
For this reason, and in complement of simulation tools, recent CNN-based methods are trying to further bridge the realism gap by learning a mapping from rendered to real data, directly in the image domain. Mostly based on unsupervised conditional generative adversarial networks (GANs)
\cite{taigman2016unsupervised,shrivastava2016learning,bousmalis2016unsupervised}
or style-transfer solutions 
\cite{gatys2016image}, 
these methods still need a set of real samples to learn their mapping.

Other approaches are instead focusing on adapting the recognition methods themselves, to make them more robust to domain changes. For instance, solutions like \textit{DANN} 
\cite{ganin2016domain} 
or \textit{ADDA} 
\cite{tzeng2017adversarial}
 are also using unlabeled samples from the target domain along the source data to teach the task-specific method domain-invariant features.
Considering real-world and industrial use-cases when only texture-less CAD models are provided, some researchers~\cite{sadeghi2016cad,tobin2017domain} are compensating the lack of target domain information by training their recognition algorithms on heavy image augmentations or on a randomized rendering engine. The claim is that with enough variability in the simulator, real data may appear just as another variation to the model. Considering similar applications (when no real samples nor texture information are available), our method follows the same principle, but applies it to the training of a domain-mapping function instead of the recognition networks. We demonstrate how this different approach not only improves the end accuracy, but also makes the overall solution more modular.

\par\noindent
\textbf{Regression of geometrical information:} 
As no textural information is provided for training, we apply our domain adaptation method to the mapping of real cluttered color images into the only prior domain: the geometrical representation of target objects, extracted from their CAD data.
The regression of such view-based shape information (\eg normal or depth maps) from monocular color images is not a new task in the field of computer vision, and it has been already explored by several works. 
The pioneer approaches tackled this complex mapping either by using probabilistic graphical models relying on hand-crafted features 
\cite{hoiem2005geometric,liu2011sift}, 
or by using feature matching between an RGB image and a set of RGB-D samples to find the nearest neighbors and warp them into a final result 
\cite{karsch2014depth,liu2014discrete}. 
Unsurprisingly, the latest works employ CNNs as a basis for their algorithms~\cite{eigen2015predicting,roy2016monocular,laina2016deeper}. Eigen~\etal \cite{eigen2014depth} are the first ones to apply a CNN (the popular AlexNet~\cite{krizhevsky2012imagenet}) to this problem, making predictions in a two-stage fashion: coarse prediction and refinement. This approach was further improved by additionally regressing labels and normals, with a refinement step for the final estimation~\cite{eigen2015predicting}. 

Another way of improving the quality of predicted depth or normal data is to use neural networks together with graph probabilistic models. Liu~\etal~\cite{liu2015deep} use a unified Deep Convolutional Neural Fields (DCNF) framework based on the combination of a CNN and conditional random field (CRF) to regress depth from monocular color images of various scenes. Their pipeline consists of two sub-CNNs with a common CRF loss layer, and yields detailed depth maps. Building on the previous framework, Cao~\etal~\cite{cao2017exploiting} train the DCNF model jointly for depth regression and semantic segmentation, demonstrating how joint training can improve the overall results. Similarly, Kendall~\etal~\cite{kendall2017multi} proposed a multi-task Bayesian network approach (including depth regression) which weighs multiple loss functions by considering the uncertainty of each task. Another way of efficiently combining a multi-task output was presented in~\cite{xu2018pad}, which uses so-called distillation modules to supervise and improve the output result. 
Unfortunately, all aforementioned methods require real labeled images from the target domain for their training, which is too strong a constraint for real-life scalable applications.
Our own method does build upon their conclusions 
\cite{eigen2015predicting,wang2015towards,gupta2016cross,kuga2017multi,kendall2017multi,xu2018pad},
 making use of a custom cross-modality network with advanced distillation to learn a robust mapping from noisy domains to synthetic ones.

\section{Methodology}
\label{sec:mth}

\begin{figure*}[t]
\centering
\includegraphics[width=\linewidth]{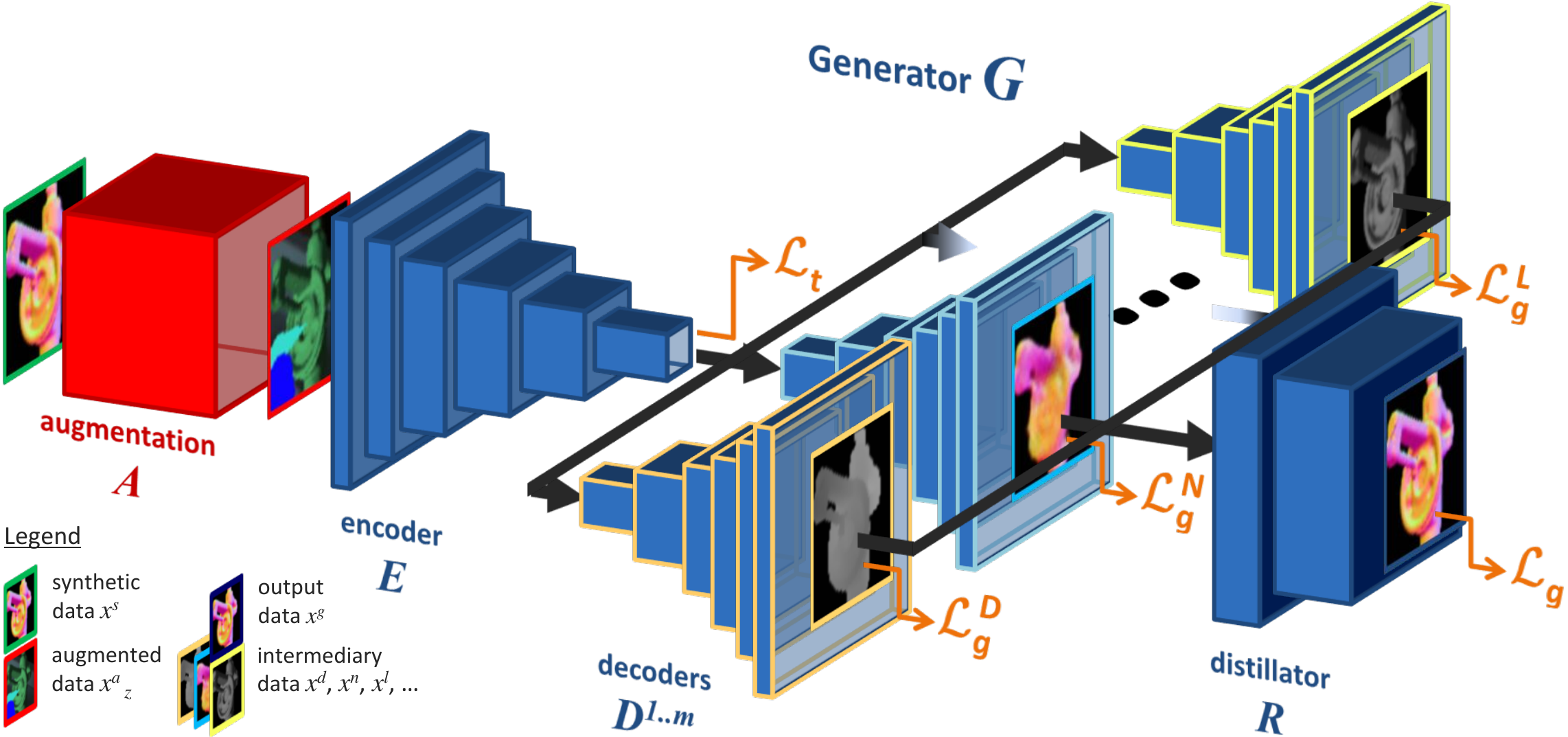}
\caption{\textbf{Training of our network $\pmb{G}$}. Taking full advantage of available synthetic data \eg texture-less CAD models, $G$ consists of a custom multi-modal pipeline with self-attentive distillation, trained to recover noiseless geometrical and semantic modalities from randomly augmented synthetic samples (detailed architecture in Figure~\ref{fig:architecture}).}
\label{fig:pipeline}  
\end{figure*}

Driven by the necessity of learning only from synthetic data for scalable recognition processes, we developed a method to map unseen real samples (\eg color images) into the noiseless synthetic domain (\eg normal maps rendered from CAD models) the task-specific solutions were trained on (\cf Figure~\ref{fig:testing}), to enable recognition.

Following the same formalization as in~\cite{zakharov2018keep}, let $X^s_c=\Set{x^s_{c,i}}{\forall i \in N^s_c}$ be a dataset made of a number $N^s_c$ of uncluttered, noiseless training samples $x^s_c$ of class $c$. 
Let $X^s=\Set{X^s_c}{\forall c \in C}$ be the complete clean training dataset. We similarly define $X^r$ the set of target $C$-related real data, completely unavailable for training. Note that samples $x^r$ can also be of a different modality than $x^s$ (\eg $x^r$ being color images while $x^s$ being normal maps, when no texture were available to render synthetic color images).
Finally, let \hbox{$T(x\ ; \theta_T)\to \widetilde{y}$} be any recognition algorithm which given a sample $x$ returns an estimate $\widetilde{y}$ of a task-specific label or feature $y$ (\eg class, pose, mask image, hash vector, \etc). We define as $T^s$ the method trained on noiseless $X^s$.

Given this setup, our pipeline trains a function $G$ purely on synthetic data (and thus in an unsupervised manner), to learn a mapping from complex $C$-related instances to their corresponding clean signal (\cf Figure~\ref{fig:pipeline}). To achieve this when no domain-relevant data is available for training, we describe in this section how $G$ is trained against a data augmentation pipeline \hbox{$A(x^s, z) \to x^a_z$}, with $z$ a noise vector randomly defined at every training iteration and $x^a_z$ the resulting noisy data. 
Our training approach assumes that $G$ removes the artificially introduced noise $z$ such that only the original synthetic signals $x^s$ are retained. Thus, $G$ can be seen as a noise filter that removes unneeded elements in input data, and can be also applied over the domain $X^r$ of real samples as long as synthetic information can be extracted from them. We demonstrate that, in the case of CAD-based visual recognition, we can indeed define a new generative method $G$ fully utilizing the synthetic modalities, and a complex and stochastic augmentation pipeline $A$ to train $G$ against, such that $G$ maps real images into the learnt synthetic domain with high accuracy. We even demonstrate how this process increases the probability that $T^s(G(x^r)) = \widetilde{y^g}$ is accurate compared to $T^a(x^r) = \widetilde{y^r}$, with $T^a$ the task-specific algorithm directly trained on data augmented by $A$.
Though we focus the rest of the paper on CAD-based visual recognition for clarity, the principles behind our solution can be directly applied to other use-cases.

\subsection{Real-to-Synthetic Mapping through Multi-Modal Distillation} \label{sec:generator}

\begin{figure*}[t]
\centering
\includegraphics[width=\linewidth]{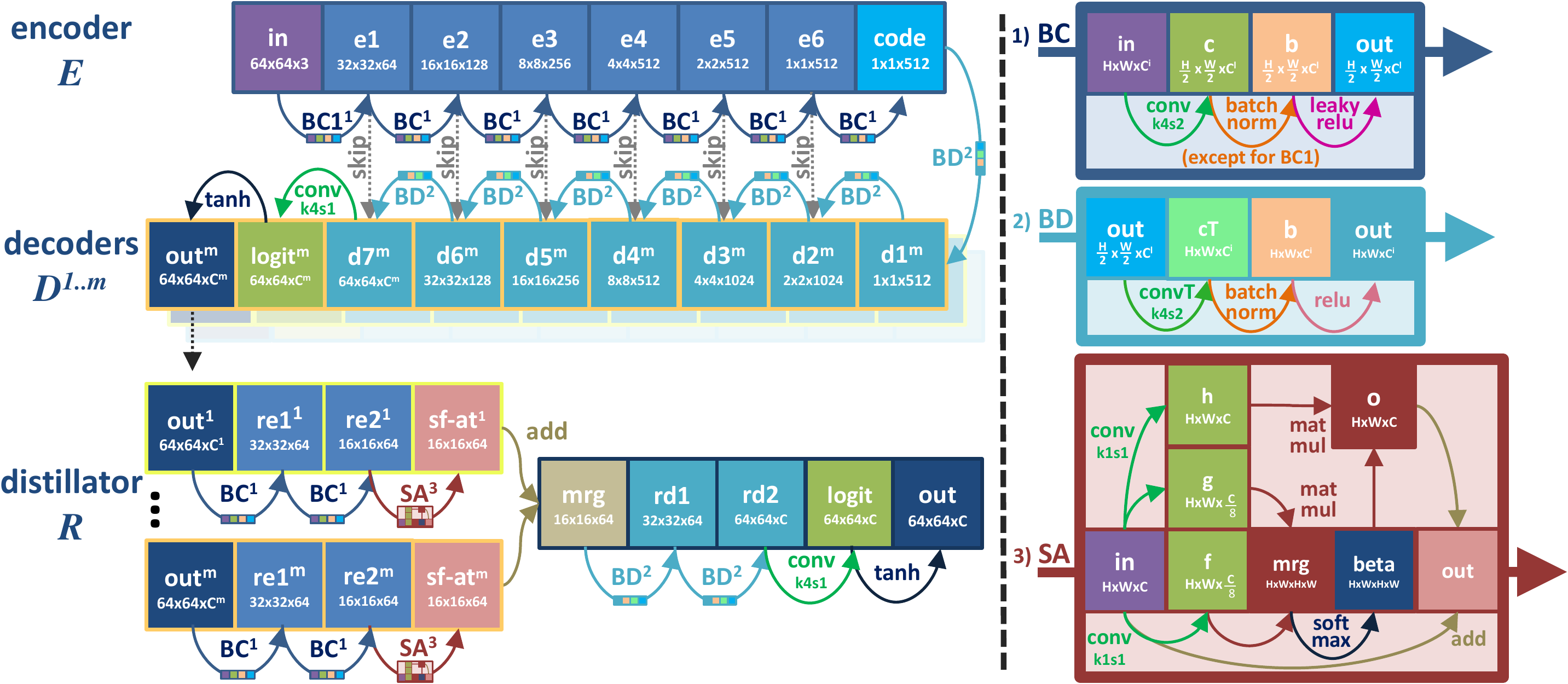}
\caption{\textbf{Detailed architecture of our network $\pmb{G}$}, on the \textit{left}. Reused layer blocks ($BC$ for encoding, $BD$  for decoding, $SA$ for self-attention) are detailed on the \textit{right}. \textit{conv k4s2} stands for a convolutional layer with $4 \times 4$ filters and a stride of 2; \textit{convT} stands for a transposed convolution.}
\label{fig:architecture}  
\end{figure*}

\subsubsection{Multi-Modal U-Net Architecture} \label{sec:unet}

As demonstrated by previous works in multi-task learning 
\cite{eigen2015predicting,wang2015towards,gupta2016cross,kuga2017multi,kendall2017multi,xu2018pad}, 
it is often advantageous to train a network on several tasks (even when considering only one), as the synergy between the tasks can make each of them perform better, and make the common layers for feature extraction more robust and focused on abstract, cross-modality information.

We thus adopt such a scheme to guide our generator in its main task of extracting the chosen synthetic features from noisy samples. Not limited to the scarce, sometimes imprecise, annotations of real training datasets, we can rely on a multitude of different synthetically-rendered modalities. For industrial CAD-based recognition, $G$ would learn to map real images into a geometrical domain (normal and/or depth maps), using for sub-tasks the regression of depth and normal maps, semantic or contour mask, \etc (\cf Section~\ref{sec:synt}).

Inspired by previous multi-modal generative pipelines~\cite{kuga2017multi,kendall2017multi,xu2018pad}, our network is composed of a single convolutional encoder $E$ and $m$ decoders $D^{mod}$, with $m$ the number of sub-tasks. For the rest of the paper, we only consider up to 4 sub-tasks---normal and depth regression, semantic segmentation, foreground lightness evaluation---though it would be straightforward to add more (\eg contour extraction as in~\cite{xu2018pad}).

In our solution, each intermediary modality is fully decoded in order to be compared to its synthetic ground-truth. 
Each generative loss $\mathcal{L}_{g}^{mod}$ (L1 distance for images, cross-entropy for binary masks) is back-propagated through its decoder, then jointly through the common encoder (\cf Figure~\ref{fig:pipeline}).

A triplet loss $\mathcal{L}_{t}$ is optionally added at the network bottleneck to improve the feature distribution in the embedding space, using task-specific metrics to push apart encoded features of images from semantically-different images, while bringing together features of similar elements.
\begin{equation} \label{eq:triplet_loss}
\mathcal{L}_{t}(E) = \mkern-22mu
\sum_{(x_{b},x_{p},x_{n}) \in X} { \mkern-22mu max\left(0,1-\frac{||E(x_{b})-E(x_{n})||_2^2}{||E(x_{b})-E(x_{p})||_2^2+m}\right)}
\end{equation}
with $x_b$ the input image used as binding anchor , $x_p$ a positive or similar sample, $x_n$ a negative or dissimilar one, and $m$ the task-specific margin setting the minimum ratio for the distance between similar and dissimilar pairs of samples. For instance, for the task of instance classification and pose estimation (ICPE), we set $m = 2 \arccos (|q_b \cdot q_p|)$ if $x_b$ and $x_p$ are images of the same class, else $m = n$ (with $q_b$ and $q_p$ the pose quaternions corresponding to $x_b$ and $x_p$, and $n > \pi$ a fixed margin).

Further distinguishing our solution from usual multi-modal auto-encoders, we add skip connections from each encoding block to its reciprocal decoding block. As demonstrated in previous works 
\cite{kuga2017multi,zhu2017unpaired}, passing high-resolution features from the contracting layers along the outputs of previous decoding blocks not only improves the training by avoiding vanishing gradients, but also guides the decoding blocks in upsampling and localizing the features. We observe a clear performance boost from this change, as shown in Table~\ref{tab:ablation}.

\subsubsection{Distillation with Self-Attention} \label{sec:distill}

\begin{figure*}[t]
\centering
\includegraphics[width=1\linewidth]{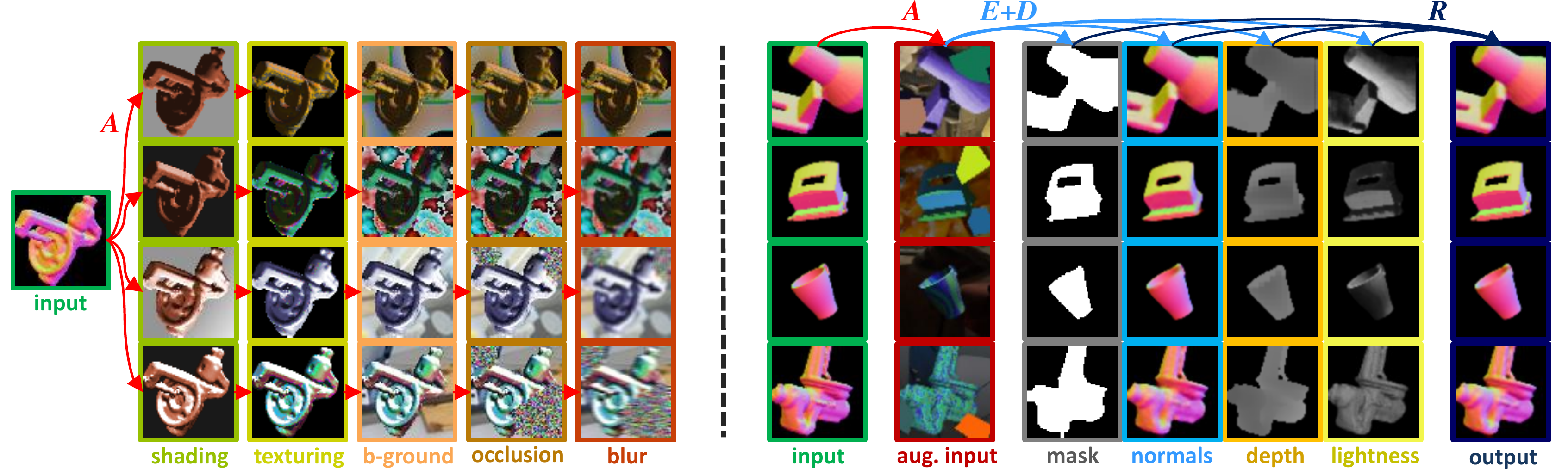}
\caption{\textbf{Augmentation and training results.} On the \textit{left}, we demonstrate how normal maps are step by step transformed into complex, random color images by our online augmentation pipeline. On the \textit{right}, we present how $G$ is trained on these images, learning to map them back to the noiseless geometrical information.}
\label{fig:augmentation_and_results}  
\end{figure*}
s
If training the target decoder along others already improves its performance by synergy, several works 
\cite{xu2018pad,gupta2016cross,neverova2017predicting} 
demonstrated how one can further take advantage of multi-modal architectures by adding a distillation module on top of the decoders, merging their outputs to distill a final result.

In their work~\cite{xu2018pad}, \textit{Pad-Net} authors present several distillation strategies, with the most efficient one making use of attention mechanism~\cite{mnih2014recurrent,ba2014multiple,luong2015effective} to better weigh the cross-modality merging, bringing forward the most relevant features for the final modality.

Using this insight, we built our own module $R$ to refine the target results from the partially re-encoded intermediary modalities by using self-attention computations 
\cite{cheng2016long}. This mechanism, adapted by Zhang~\etal~\cite{zhang2018self} for image generation and detailed in Figure~\ref{fig:architecture}, is used to efficiently model relationships between widely-separated spatial regions. Given a feature map $x \in \mathbb{R}^{C \times H \times W}$, the output of the self-attention operation is:
\begin{equation}
x_{sa} = x + \gamma \cdot \sigma\big((W_f \ast x)^\intercal \cdot (W_g \ast x)\big) \cdot (W_h \ast x)
\end{equation}
with $\sigma$ the \textit{softmax} activation function; $W_f \in \mathbb{R}^{\bar{C} \times C}$, $W_g \in \mathbb{R}^{\bar{C} \times C}$, $W_h \in \mathbb{R}^{C \times C}$ learned weight matrices (we opt for $\bar{C} = \sfrac{C}{8}$ as in~\cite{zhang2018self}); and $\gamma$ a trainable scalar weight.
Instantiating and applying this process to each re-encoded modality, we sum the resulting feature maps, before decoding them to obtain the final output. This new distillation process not only allows to pass messages between the intermediary modalities, but also between distant regions in each of them.

Our distillator is trained jointly with the rest of the generator, with a final generative loss $\mathcal{L}_{g}$ (L1 distance here) applied to the distillation results. Not only our whole generator can thus be efficiently trained in a single pass, but no manual weighing of the sub-task losses is needed, as the distillator implicitly covers it (this furthermore suits our use-cases, as manual fine-tuning is technically possible only when validation data from target domains are available).

\subsection{Learning from Purely Geometrical CAD Data} \label{sec:synt}

\subsubsection{Synthetic Data Generation} \label{sec:render}

The aforementioned architecture has been developed to especially shine for one particular use-case, poorly covered in the literature despite being common in industrial applications: the training of recognition methods on pure 3D CAD data, \ie without any real relevant images and their annotations, nor captured textures for the 3D models to render realistic images.
Despite the apparent meagerness of the available training data, covering only the geometrical aspects of target classes with no appearance information, it is still possible to render multiple synthetic modalities from the CAD models in order to build a rich annotated dataset, to guide the training of complex generative networks such as our proposed one.

Without any relevant texture information, usual dataset rendering and training methods for the color domain cannot be directly applied.
Since only geometrical information is made available, we select the surface normal and/or depth domains as target modality for the mapping performed by $G$. For this reason and similarly to other view-based training methods from CAD data 
\cite{Wohlhart15,zakharov2017,zakharov2018keep}, 
we use a simple 3D engine to generate noiseless normal and depth maps for each class $c$ from a large set of relevant viewpoints (\eg defined as vertices of an icosahedron centered on the target elements). This dataset of geometrical mappings is both used as ground-truth for the final outputs of $G$ and some of its sub-tasks, and as inputs for the augmentation pipeline $A$ deployed when no color data is available to train $G$.

\subsubsection{Online Color Rendering and Augmentation} \label{sec:augment}

\begin{figure*}[t!]
\centering
\includegraphics[width=1\linewidth]{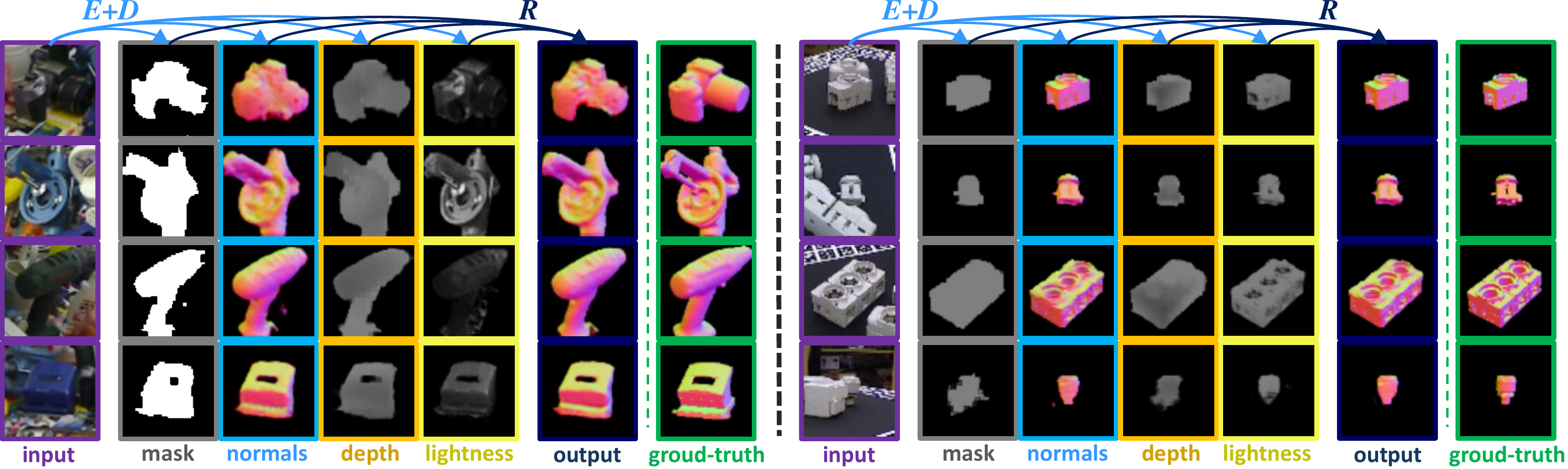}
\caption{\textbf{Qualitative results} (intermediary and final mappings), when applying $G$ purely trained on synthetic data to real samples, on LineMOD~\cite{hinterstoisser2012model} and T-LESS~\cite{hodan2017t} datasets.}
\label{fig:visual}  
\end{figure*}

$A(x^s, z) \to x^a_z$ is an extensive online augmentation pipeline we designed, parametrized by a noise vector $z$ randomly sampled at every call from a $k$-dimensional finite set $\mathbb{Z}^k$, with $k$ the number of augmentation parameters.
In order to make up for the complete lack of appearance and clutter information, $A$ follows the principle of \textit{domain randomization} 
\cite{tobin2017domain}, 
\ie it is meant to add enough visual variability to the training inputs so that the trained method can generalize to real unseen samples.
This means conceiving an augmentation pipeline with a large enough $|\mathbb{Z}^k|$. In our case, $A$ first dynamically transforms the input geometrical views into color images through random shading and texturing, before applying further noise and clutter to the images, in order to prepare $G$ for the complexity of real data.

To maximize the training variability, $A$ is built to run in parallel of GPU-based trainings (\textit{online}), providing new randomized samples every iteration (unlike \textit{offline} solutions, generating a fixed training dataset beforehand).
Inspired by the literature both in computer vision 
\cite{ounsworth2015anticipatory} 
and computer graphics 
\cite{blinn1977models,perlin2002improving},
the following operations are thus composing $A$ (illustrated in Figure~\ref{fig:augmentation_and_results}):

\par\noindent
\textbf{Simple random shading:}
$A$ first takes the provided normal maps and convert them into color images by applying simple Blinn-Phong shading 
\cite{blinn1977models}. Randomly sampling ambient and directional light sources, as well as diffusion and specular color factors for the objects, the provided surface normals are used to compute the diffuse and specular lightness maps through direct matrix products. Since distance information is lost in normal maps, this shading model is simplified by supposing the light sources at an infinite distance, hence the same light source vector for every surface point. This way, one can easily simulate an infinity of lighting conditions, returning the resulting lightness map. 

\par\noindent
\textbf{Stochastic texturing:}
Given the lack of relevant texture information, random texture maps are procedurally generated using noise functions \eg fractal Perlin noise~\cite{perlin2002improving} and Voronoi texturing.
Downsampled to 2D vector maps, the original normals are used to index the generated textures; to achieve a more ``organic" appearance, with patterns sometimes following some of the shape features.

\par\noindent
\textbf{Background addition:}
To simulate cluttered scenes, backgrounds are added to the rendered images, either re-using the previously-introduced noise functions, or using random patches from any publicly available image dataset (\eg COCO~\cite{lin2014microsoft}). Lightness maps from the shading step are furthermore used to homogenize the background brightness.

\par\noindent
\textbf{Random occlusion:}
Occlusions are introduced to further simulate clutter, but also so that $G$ can learn to recover hidden or lost geometrical information.
Based on~\cite{ounsworth2015anticipatory}, occluding polygons are generated by walking around the image, taking random angular steps and random radii at each step; then by painting it on top of the images with color noise.

\par\noindent
\textbf{Blur:}
To reproduce possible motion blur or unfocused images, Gaussian, uniform or median blur is applied with variable intensity.

\section{Evaluation}
\label{sec:exp}

Pursuing our application of \textit{SynDA} to CAD-based recognition tasks, we evaluate our method on two different tasks of localized object classification and pose estimation, opting for well-known algorithms on datasets commonly used in this domain~\cite{Wohlhart15,bousmalis2016unsupervised,zakharov2018keep}. First presenting concise qualitative observations, we then quantitatively and extensively evaluate \textit{SynDA} through a comparison of its performance to state-of-the-art solutions depending on the available training modalities, and through an ablation study.

\subsection{Experimental Setup} \label{sec:tasks}

\begin{table*}[t]
\centering
\caption{
\textbf{Quantitative comparison of recognition pipelines}, depending on the available training data, for the task of localized instance classification on T-LESS~\cite{hodan2017t} with $T$ ResNet9 network~\cite{he2016identity}. \textit{Methods are explicitly described in annex}.}
\label{tab:modalities_tless} 
\resizebox{1\linewidth}{!}{
\def\arraystretch{1}
\begin{scriptsize}
\begin{tabu}{@{}l@{\hskip 15pt}l@{\hskip 10pt}c@{\hskip 20pt}c@{}}
\toprule
\multicolumn{3}{l@{\hskip 20pt}}{\textbf{Training}} & 
\multirow{2}{*}{\shortstack{\textbf{Classification}\\\textbf{accuracy}}} \\ 
\cmidrule(lr){1-3} 
\rule[5pt]{0pt}{5pt}
\textbf{Data} & \textbf{Method} & & \\

\midrule

\parbox[c]{8em}{
\includegraphics[width=\linewidth]{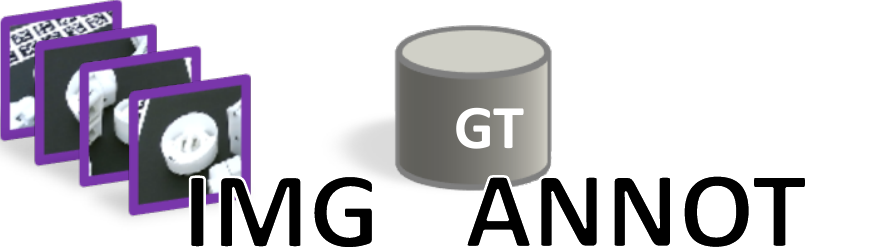}
}
& $T(X_{train}^r)\to y$ & ($\varnothing$) & \textbf{99.34\%} \\
\midrule
\multirow{2}{*}{\parbox[c]{8em}{
\includegraphics[width=\linewidth]{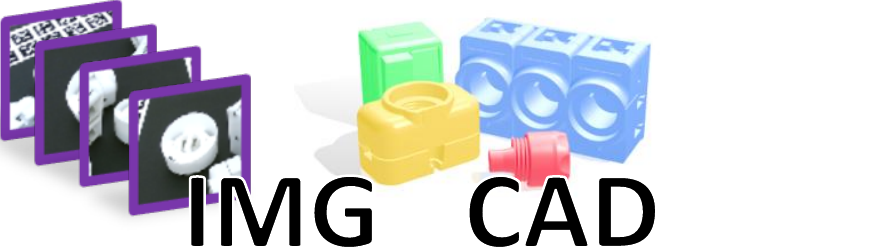}
}}	
& $T^d\big(A(X^{s})\ ;\ X_{train}^r\big)\to y, c_{dom}$ & (DANN) & 60.58\% \\
& $G^{pix}(X^s, X_{train}^r) \to X^{r'}\ ,\ T(X^{r'})\to y$ & (PixelDA) & 63.12\% \\
\midrule

\parbox[c]{8em}{
\includegraphics[width=\linewidth]{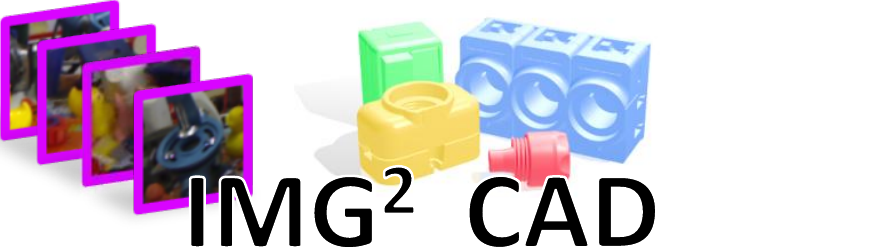}
}	
& $G^{pre}(X_{other}^r) \to X^s \ ,\ T(X^s)\to y$ & (Iro~\etal)& 36.03\% \\
\midrule

\multirow{2}{*}{\parbox[c]{8em}{
\includegraphics[width=\linewidth]{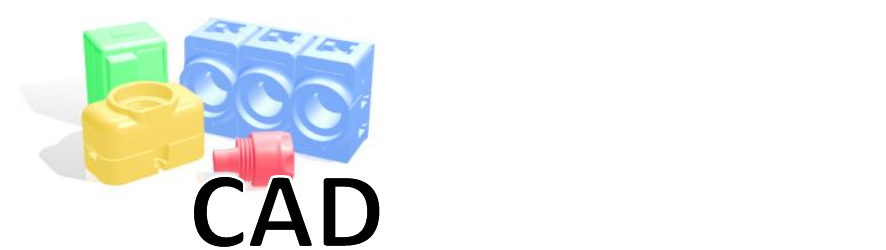}
}}	

& $T\big(A(X^s)\big)\to y$ & ($\varnothing$) & 53.81\% \\
\rowfont{\color{blue}}
& $G(A(X^s)\big) \to X^s \ ,\ T(X^s)\to y$  & (ours) & 71.78\% \\
\bottomrule
\end{tabu}
\end{scriptsize}
}  
\end{table*}

\subsubsection{Instance Classification (IC) on T-LESS} \label{sec:tless}

As a preliminary experiment, we consider localized classification on T-LESS~\cite{hodan2017t}, a dataset of industrial objects with texture-less CAD models and RGB-D images from different complex scenes. Strong textural and geometrical similarities between the objects and heavy occlusions make it a challenging dataset for geometry-based classification. As in~\cite{zakharov2018keep}, we select the first 3 scenes and their 11 objects, building a set of 5,514 RGB patches of objects occluded up to 60\%.
For our task-specific method, we opt for the well-known ResNet 
\cite{he2016identity}, with 9 residual blocks.

\subsubsection{Instance Classification and Pose Estimation (ICPE) on LineMOD} \label{sec:linemod}

LineMOD~\cite{hinterstoisser2012model} contains 15 mesh models of distinctive textured objects, along their RGB-D sequences and camera poses. We take advantage of this dataset to demonstrate how texture information is too often taken for granted in CAD-based application, and how its absence can heavily impact usual methods (\eg in industrial settings). LineMOD has 4 symmetric objects. While some works simply removed these ambiguous elements for evaluation~\cite{bousmalis2016unsupervised,bousmalis2016domain}, others only constrain the real views by keeping the unambiguous poses for these 4 objects~\cite{Wohlhart15,zakharov2017,zakharov2018keep}. We opt for the latter solution, to highlight the generalization capabilities of \textit{SynDA} \wrt the number of objects.
To further demonstrate that our method is tailored neither to a dataset nor to a recognition method, we select a different solution, the so-called \textit{triplet} CNN~\cite{Wohlhart15,zakharov2017}, which uses the aforementioned triplet loss $\mathcal{L}_{t}$, to map images to an embedding space which enforces separation for distinct classes and poses.

\begin{table*}[t]
\centering
\caption{
\textbf{Quantitative comparison of recognition pipelines}, depending on the available training data, for the task of localized instance classification and pose estimation (ICPE) on LineMOD~\cite{hinterstoisser2012model} with $T$ triplet CNN~\cite{Wohlhart15,zakharov2017}.}
\label{tab:modalities} 
\resizebox{1\linewidth}{!}{
\def\arraystretch{1}
\begin{tabu}{@{}l@{\hskip 10pt}lc@{\hskip 10pt}c@{\hskip 10pt}cc@{}}
\toprule
\multicolumn{3}{l}{\textbf{Training}} & \multicolumn{2}{c}{\textbf{Angular error}} & \multirow{2}{*}{\shortstack{\textbf{Classification}\\\textbf{accuracy}}} \\ \cmidrule(lr){1-3} \cmidrule(lr){4-5}
\rule[5pt]{0pt}{5pt}
\textbf{Data} & \textbf{Method} & & {\hskip 5pt}\textbf{Median} & \textbf{Mean} &  \\

\midrule

\parbox[c]{8em}{
\includegraphics[width=\linewidth]{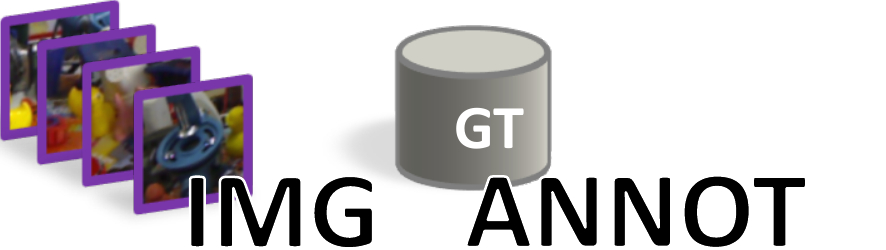}
}
& $T(X_{train}^r)\to y$ & ($\varnothing$) &{\hskip 5pt} \textbf{9.50$\pmb{^{\circ}}$}& \textbf{12.42$\pmb{^{\circ}}$} & \textbf{99.72\%} \\
\midrule

\multirow{2}{*}{\parbox[c]{8em}{
\includegraphics[width=\linewidth]{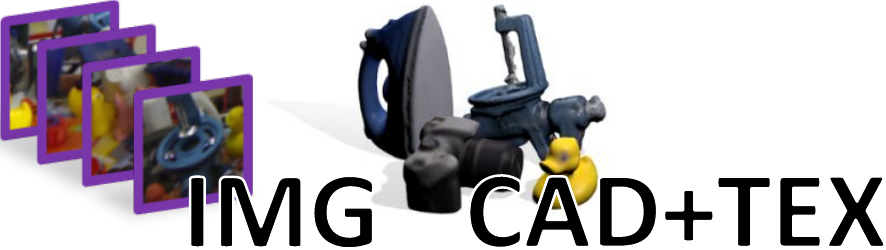}
}}		
& $T^d\big(A(X^{s,t})\ ;\ X_{train}^r\big)\to y, c_{dom}$ & (DANN) & {\hskip 5pt}14.33$^{\circ}$ & 30.45$^{\circ}$ & 89.84\% \\
& $G^{pix}(X^{s,t}, X_{train}^r) \to X^{r'}\ ,\ T(X^{r'})\to y$ & (PixelDA) & {\hskip 5pt}15.38$^{\circ}$ & 35.17$^{\circ}$ & 91.06\% \\
\midrule

\multirow{2}{*}{\parbox[c]{8em}{
\includegraphics[width=\linewidth]{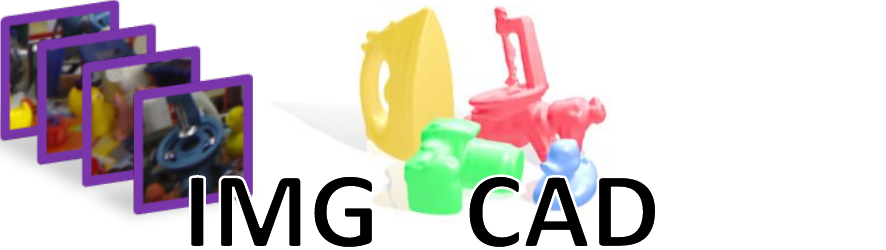}
}}	
& $T^d\big(A(X^{s})\ ;\ X_{train}^r\big)\to y, c_{dom}$ & (DANN) & {\hskip 5pt}43.63$^{\circ}$ & 68.59$^{\circ}$ & 40.13\% \\
& $G^{pix}(X^s, X_{train}^r) \to X^{r'}\ ,\ T(X^{r'})\to y$ & (PixelDA) & {\hskip 5pt}95.14$^{\circ}$ & 97.36$^{\circ}$ & 35.39\% \\
\midrule

\multirow{2}{*}{\parbox[c]{8em}{
\includegraphics[width=\linewidth]{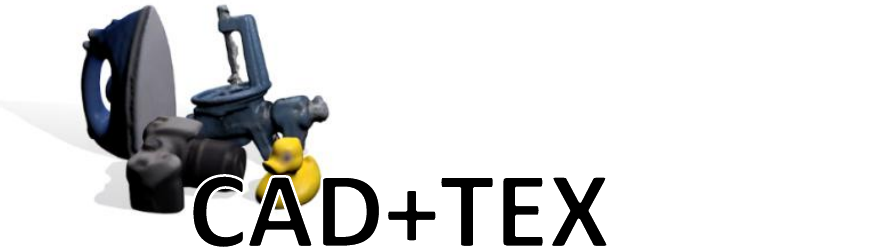}
}}		
& $T(X^{s,t})\to y$ & ($\varnothing$) & {\hskip 5pt}88.62$^{\circ}$ & 92.35$^{\circ}$ & 43.62\% \\
& $T\big(A(X^{s,t})\big)\to y$ & ($\varnothing$) & {\hskip 5pt}70.18$^{\circ}$ & 84.22$^{\circ}$ & 49.11\% \\
\midrule

\parbox[c]{8em}{
\includegraphics[width=\linewidth]{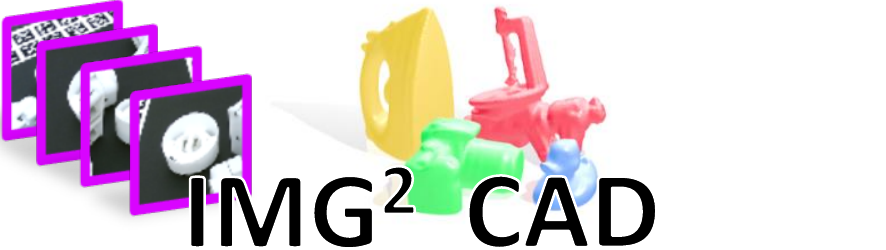}
}	
& $G^{pre}(X_{other}^r) \to X^s \ ,\ T(X^s)\to y$ & (Iro~\etal)& {\hskip 5pt}52.43$^{\circ}$ & 71.69$^{\circ}$ & 41.49\% \\
\midrule

\multirow{2}{*}{\parbox[c]{8em}{
\includegraphics[width=\linewidth]{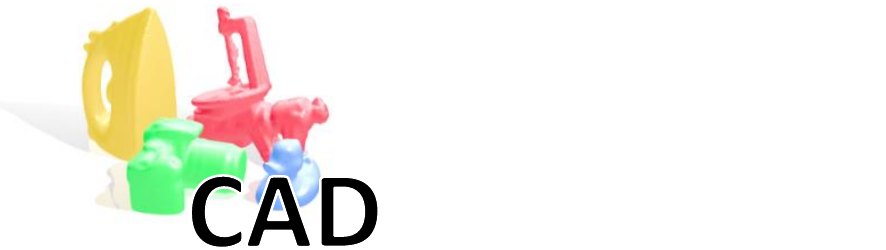}
}}	

& $T\big(A(X^s)\big)\to y$ & ($\varnothing$) & {\hskip 5pt}41.23$^{\circ}$ & 67.50$^{\circ}$ & 34.38\% \\
\rowfont{\color{blue}}
& $G(A(X^s)\big) \to X^s \ ,\ T(X^s)\to y$  & (ours) & {\hskip 5pt}13.37$^{\circ}$ & 27.46$^{\circ}$ & 91.28\% \\
\bottomrule
\end{tabu}
}
\end{table*}

\subsubsection{Tasks Preparation} \label{sec:preparation}

For both tasks, synthetic training data $X^s$ (normal, depth and semantic maps) are generated with a basic 3D engine (while lightness maps are obtained from $A$), following the procedure described in Section~\ref{sec:synt}, taking into account in-plane rotations. For each task, the real color images are split 50/50 into a test set $X^r_{test}$, and a training set $X^r_{train}$ for comparative methods which require real data. Similarly, we render $X^{s,t}$, realistic images from textured models to train some opponent methods on.

\subsection{Qualitative Observations} \label{sec:qualita}

Qualitative results can be found in Figures~\ref{fig:testing}, \ref{fig:augmentation_and_results}, \ref{fig:visual}, as well as in the supplementary material. 
For both datasets, our method clearly learns to recover the clean geometrical features of target objects in unseen real images (Fig.~\ref{fig:visual}), even though it has been trained with no information about the real domain (Fig.~\ref{fig:augmentation_and_results}).
The monochrome appearance of T-LESS objects may make the task easier; but as this information is not known during training, $G$ is still trained on random noisy textures, and yet manages to map the real samples and even to recover occluded parts. As demonstrated on LineMOD, our solution indeed learns to ignore visual properties such as textures to retain the synthetic features, using the prior CAD data.

GAN-based domain adaptation methods such as \textit{PixelDA}~\cite{bousmalis2016unsupervised} fail to learn their opposite mapping when only geometrical properties are provided, as shown in the annex. Indeed, learning both to add clutter and assign the proper texture to each class is a much more complex task, which would require further supervision (for instance, the foreground-similarity loss of \textit{PixelDA} cannot be used in this setting to guide the network).

Finally, we can visually observe the improvements between the intermediary normal maps (directly from decoder $D^N$) and the refined outputs after self-attentive distillation, both in terms of segmentation and internal details. As mentioned in Section~\ref{sec:unet}, one could easily add or replace intermediary modalities (for instance, regressing the objects lightness maps may not seem fully relevant, though it can be used to provide the latest layers of the network with information from the original color domain).

\subsection{Comparison with other Domain Adaptation Approaches}

Given the two pre-defined evaluation tasks, we quantitatively evaluate the performance of our pipeline, and compare it with usual, state-of-the-art methods, depending on the available training data (real images, corresponding annotations, CAD models, corresponding realistic textures, or real images from a different domain). For each setup, the same task-specific network is used (ResNet for IC on T-LESS, Triplet CNN for ICPE on LineMOD), trained by itself, against our augmentation pipeline $A$ (with texturing augmentation disabled for pre-textured data), or along some auxiliary generators or sub-networks for domain adaptation (\eg for \textit{PixelDA}~\cite{bousmalis2016unsupervised} or DANN~\cite{ganin2016domain}; for $T$ used with a pre-trained monocular-RGB-to-depth generator $G^{pre}$~\cite{laina2016deeper}; or for \textit{SynDA}).

For both tasks, we consistently observe the positive impact of \textit{SynDA} on recognition, as shown in Tables~\ref{tab:modalities_tless}-\ref{tab:modalities}. 
Despite being trained on the scarcest data, with the largest domain gap, our generator $G$ brings the performance of the task-specific methods $T$ above other solutions trained on more relevant information. The accuracy improvement is even more apparent for the pose regression task, as our pipeline precisely recovers geometrical features.
It also appears clear that decoupling data augmentation and recognition training is beneficial, as illustrated by the accuracy difference between the two last lines of each table. This follows our initial intuition on the logic of teaching task methods in the available clean synthetic domain, while learning in parallel a mapping to project real data into this prior domain. This separation furthermore makes it straightforward to train new task-specific methods, with $G$ ready to be plugged on top.

\subsection{Architecture Validation through Ablation}

\begin{table*}[t]
\centering
\caption{
\textbf{Architectural ablation study}, with the ``IC on LineMOD" task.}
\label{tab:ablation} 
\resizebox{1\linewidth}{!}{
\def\arraystretch{1}
\begin{tabu}{@{}c|cccc|c|c@{\hskip 10pt}c|c@{\hskip 5pt}c|c@{\hskip 10pt}c@{\hskip 10pt}c@{}}
\toprule
\multicolumn{1}{c}{\textbf{Encoder}} & \multicolumn{4}{c}{\textbf{Decoders}} & \multicolumn{1}{c}{\textbf{Distill.}} &  \multicolumn{2}{c}{\textbf{Layers}} &  \multicolumn{2}{c@{\hskip 5pt}}{\textbf{Losses}}  & \multicolumn{2}{c}{\textbf{Angular error}} & \multirow{2}{*}{\shortstack{\textbf{Classification}\\\textbf{accuracy}}} \\
\cmidrule(lr){0-0}\cmidrule(lr){2-5}\cmidrule(lr){6-6}\cmidrule(lr){7-8}\cmidrule(lr){9-10}
\cmidrule(lr){11-12}
{\color{archi_e}\textbf{$E$}} & {\color{archi_dn}\textbf{$D^N$}} & {\color{archi_dd}\textbf{$D^D$}} & {\color{archi_dm}\textbf{$D^M$}} & {\color{archi_dl}\textbf{$D^L$}} & {\color{archi_e}\textbf{$R$}} & {\color{archi_sa}\textbf{$\overrightarrow{SA}$}} & {\color{archi_dm}\textbf{$\overrightarrow{skip}$}} & {\color{archi_loss}\textbf{$\mathcal{L}_{g}^{1..m}$}} & {\color{archi_loss}\textbf{$\mathcal{L}_{t}$}}
&{\hskip 5pt} \textbf{Median} & \textbf{Mean} &  \\

\midrule

$\checkmark$ & $\checkmark$ & & & & & & $\checkmark$ & $\checkmark$ &  & {\hskip 5pt}15.75$^{\circ}$ & 32.80$^{\circ}$ & 87.35\% \\
\midrule
$\checkmark$ & $\checkmark$ & & & & & & $\checkmark$ & $\checkmark$ & $\checkmark$ & {\hskip 5pt}15.76$^{\circ}$ & 33.76$^{\circ}$ & 88.04\% \\
\midrule
$\checkmark$ & $\checkmark$ & $\checkmark$ & & & $\checkmark$ & $\checkmark$ & $\checkmark$ & $\checkmark$ &  $\checkmark$ & {\hskip 5pt}14.32$^{\circ}$ & 30.31$^{\circ}$ & 89.00\% \\	
\midrule
$\checkmark$ & $\checkmark$ & & $\checkmark$ & & $\checkmark$ & $\checkmark$ & $\checkmark$ & $\checkmark$ & $\checkmark$ & {\hskip 5pt}14.48$^{\circ}$ & 30.71$^{\circ}$ & 89.32\% \\	
\midrule
$\checkmark$ & $\checkmark$ & $\checkmark$ & $\checkmark$ & & $\checkmark$ & $\checkmark$ & $\checkmark$ & $\checkmark$ & $\checkmark$ & {\hskip 5pt}14.22$^{\circ}$ & 29.26$^{\circ}$ & 89.67\% \\	
\midrule
$\checkmark$ & $\checkmark$ & $\checkmark$ & $\checkmark$ & $\checkmark$ & & & $\checkmark$ & $\checkmark$ & $\checkmark$ & {\hskip 5pt}14.66$^{\circ}$ & 30.83$^{\circ}$ & 88.59\% \\		
\midrule
$\checkmark$ & $\checkmark$ & $\checkmark$ & $\checkmark$ & $\checkmark$ & $\checkmark$ & & & $\checkmark$ & $\checkmark$ & {\hskip 5pt}16.07$^{\circ}$ & 33.22$^{\circ}$ & 87.69\% \\
\midrule
$\checkmark$ & $\checkmark$ & $\checkmark$ & $\checkmark$ & $\checkmark$ & $\checkmark$ & & $\checkmark$ & $\checkmark$ & $\checkmark$ & {\hskip 5pt}14.43$^{\circ}$ & 29.56$^{\circ}$ & 90.38\% \\	
\midrule
$\checkmark$ & $\checkmark$ & $\checkmark$ & $\checkmark$ & $\checkmark$ & $\checkmark$ & $\checkmark$ & $\checkmark$ & $\checkmark$ & $\checkmark$ & {\hskip 5pt}\textbf{13.37$\pmb{^{\circ}}$} & \textbf{27.46$\pmb{^{\circ}}$} & \textbf{91.28\%} \\	
\bottomrule
\end{tabu}
}
\end{table*}

Table~\ref{tab:ablation} presents the results of an extensive ablation study done on our novel network architecture. By consolidating several state-of-the-art works on generative networks~\cite{kuga2017multi,kendall2017multi,xu2018pad,zhang2018self}, we developed a robust architecture to tackle extreme domain mappings (\eg real RGB to synthetic normals).

As mentioned in Section~\ref{sec:qualita}, we can observe how the addition of decoders for auxiliary tasks improves the final output by synergy. The inclusion of self-attention mechanism ($SA$ layers) in the distillation module further enhances this effect, weighting the contribution of features between intermediary modalities, but also between distant internal regions.
Finally, the benefits of passing messages directly between each encoder block and their opposite block for each decoder $D^{1..m}$, through the use of $skip$ layers (\cf U-Net architectures 
\cite{kuga2017multi,zhu2017unpaired}),
is clearly highlighted in the table, as well as the use of a triplet loss $\mathcal{L}_{t}$ at the bottleneck to improve the quality of the embedding space.

All in all, our network relies on a powerful multi-task architecture, structured to tackle real-to-synthetic mapping challenges, by utilizing any available synthetic modalities to learn robust features. One could easily build on this solution by considering additional or more use-case relevant sub-tasks (\eg contour regression, part segmentation, \etc).

\section{Conclusion}
\label{sec:cnc}

We present \textit{SynDA}, a novel strategy for complex domain adaptation scenarios. Applied to several CAD-based recognition tasks and making use of a state-of-the-art generative network, our solution outperforms other supervised or unsupervised methods. For the challenging task of localized instance recognition and pose estimation \eg on RGB LineMOD data, \textit{SynDA} more than doubles the angular and class accuracy compared to other methods trained on synthetic data, and even surpasses previous domain adaptation methods requiring real data.

This is made possible by tackling the domain mapping from the opposite direction, using our custom generator to denoise unseen real samples and retain only the recognition-relevant features available during training.


\newpage

\appendix
\beginsupplement

\noindent
{\Large\textbf{Supplementary Material}}


\begin{figure*}[b!]
\centering
\includegraphics[width=1\linewidth]{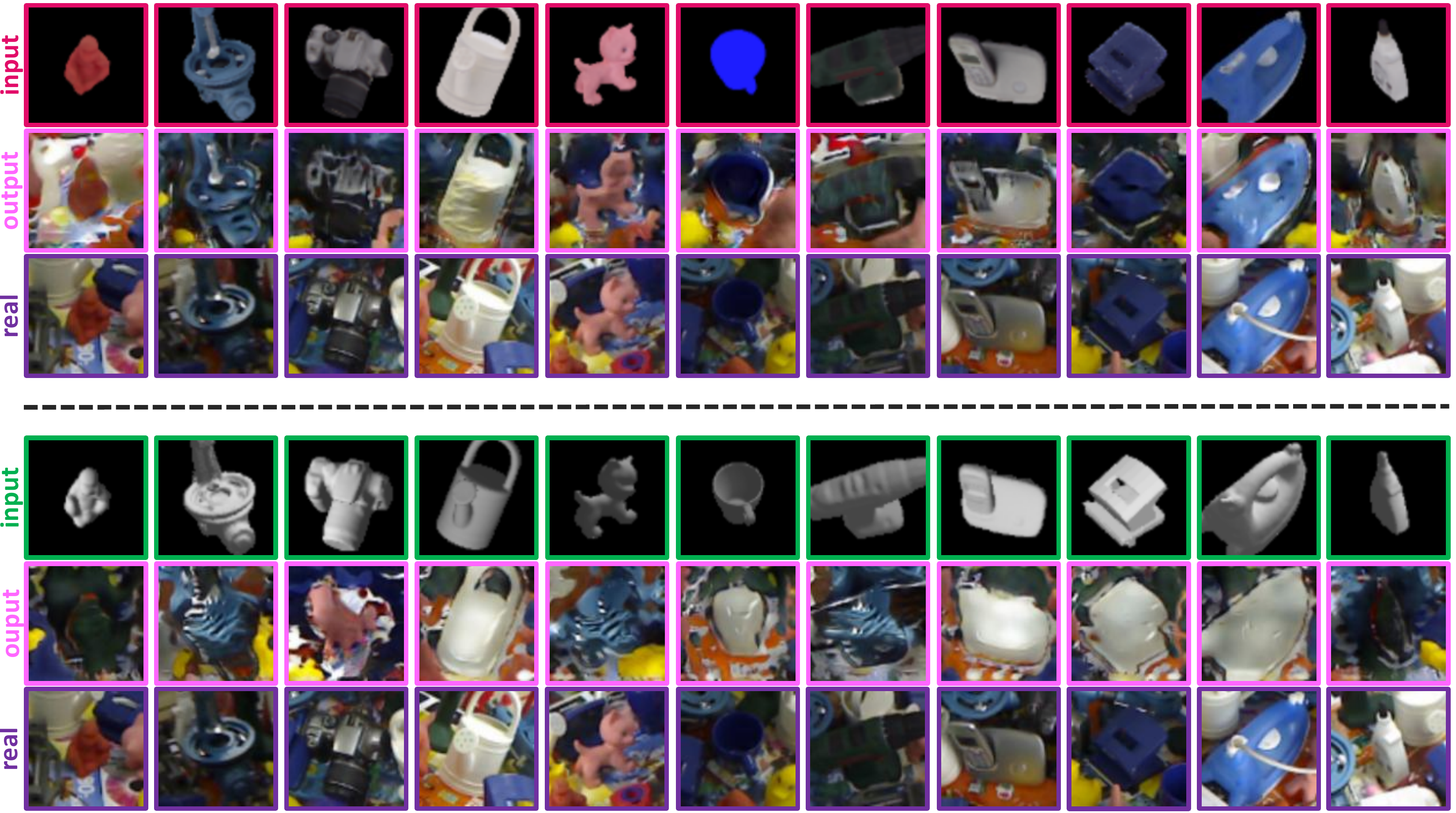}
\caption{\textbf{Qualitative comparison of results for \textit{PixelDA}~\cite{bousmalis2016unsupervised} trained with or without realistic texturing of the target objects}. The method fails to bridge the realism gap when too wide.}
\label{fig:pixelda}  
\end{figure*}

\begin{table*}[!htbp]
\centering
\caption{
\textbf{Visual comparison of recognition schemes}, depending on the available training data.}
\label{tab:comp} 
\resizebox{0.9\linewidth}{!}{
\def\arraystretch{1}
\begin{tabu}{@{}lc@{\hskip 5pt}l@{\hskip 5pt}@{\hskip 5pt}c@{}}  
\toprule
\multicolumn{3}{l@{\hskip 5pt}}{\textbf{Training}} & \multirow{2}{*}{\textbf{Inference}}
\\
\cmidrule(lr){1-3}
\rule[5pt]{0pt}{5pt}
\textbf{Data} & \multicolumn{2}{l@{\hskip 5pt}}{\textbf{Method}} &  \\
\midrule

\parbox[c]{8em}{
\includegraphics[width=\linewidth]{symbol_real+ann2}
}
& supervised &
\parbox[c]{6.61417em}{\includegraphics[width=\linewidth]{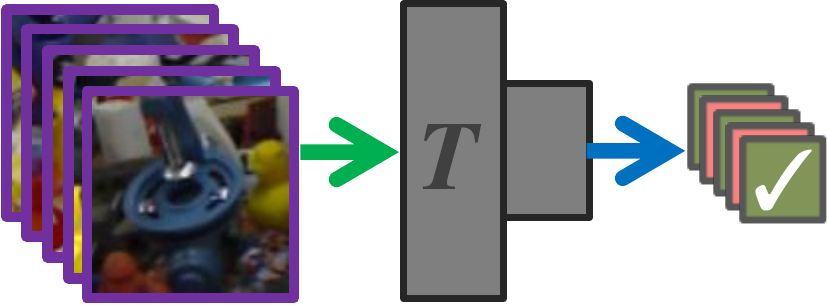}
}
& 
\parbox[c]{6.61417em}{\includegraphics[width=\linewidth]{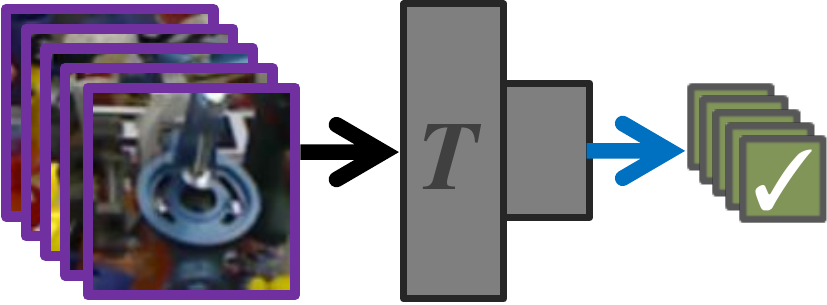}
}
\\
\midrule

\multirow{2}{*}{\parbox[c]{8em}{
\includegraphics[width=\linewidth]{symbol_real+cad+tex2}
}
}	 

& \textit{DANN}~\cite{ganin2016domain} &
\parbox[c]{10em}{\includegraphics[width=\linewidth]{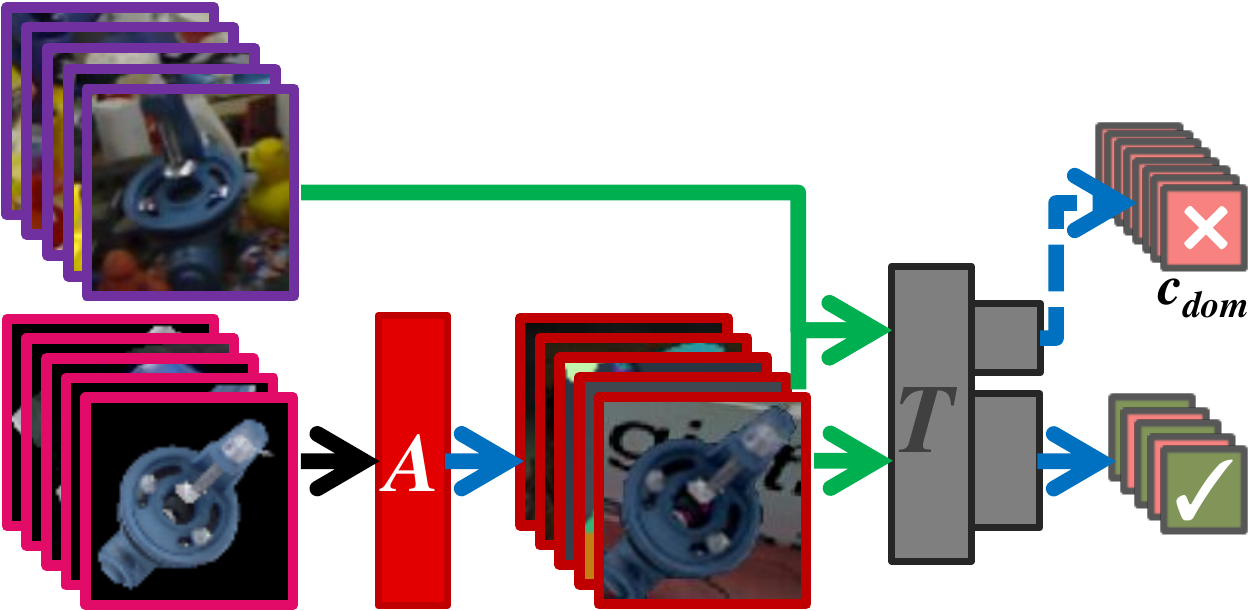}
}
&
\parbox[c]{6.61417em}{\includegraphics[width=\linewidth]{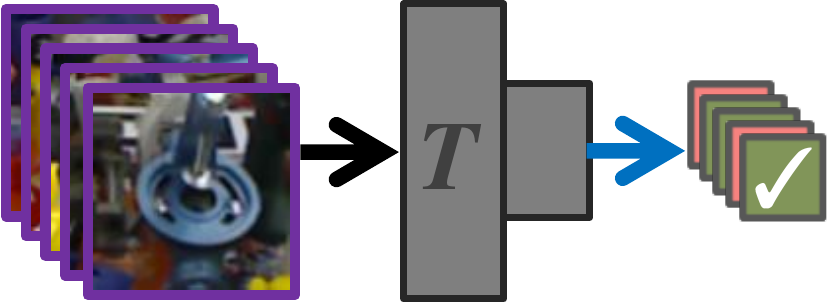}
}
\\[2pt]
\cline{2-4}  \rule[5pt]{0pt}{23pt}
& \textit{PixelDA}~\cite{bousmalis2016unsupervised} &
\parbox[c]{10em}{\includegraphics[width=\linewidth]{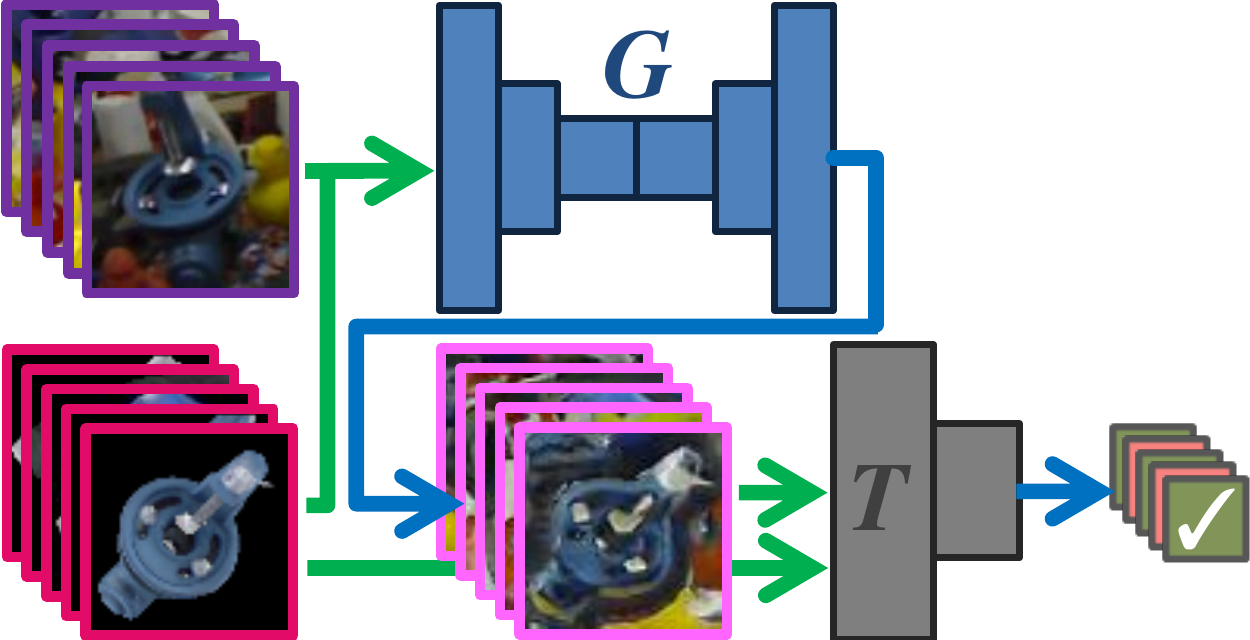}
}
&
\parbox[c]{6.61417em}{\includegraphics[width=\linewidth]{methods/real_rgb+tex_test_pixelda}
}
\\
\midrule

\multirow{2}{*}{\parbox[c]{8em}{
\includegraphics[width=\linewidth]{symbol_real+cad2}}
}	
& \textit{DANN}~\cite{ganin2016domain} &
\parbox[c]{10em}{\includegraphics[width=\linewidth]{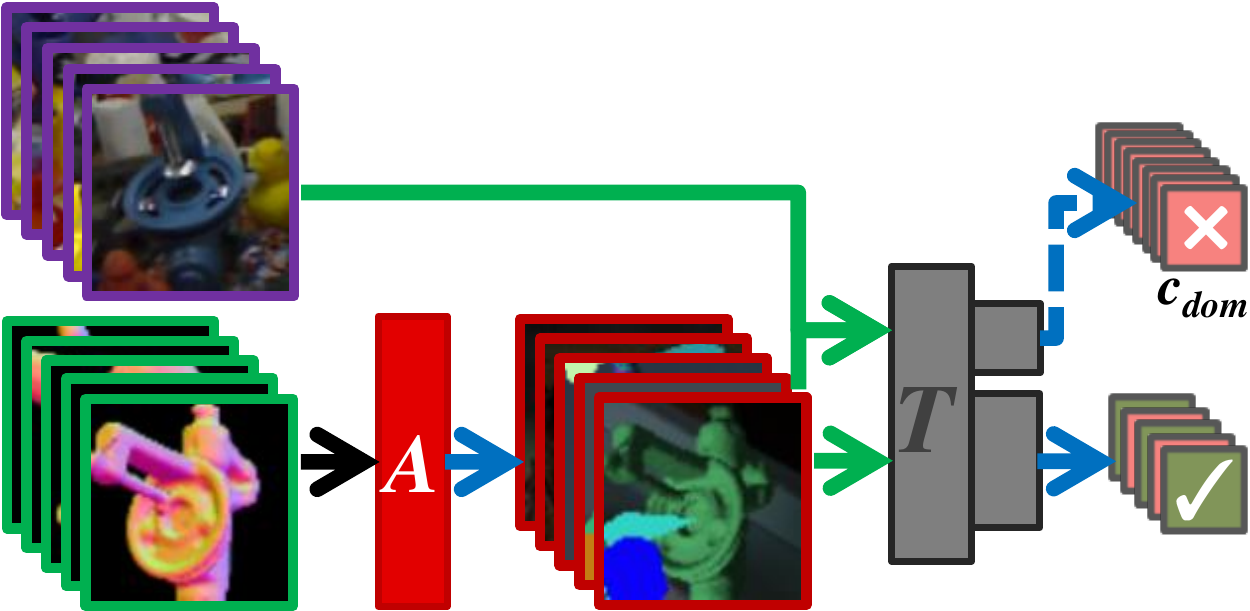}
}
&
\parbox[c]{6.61417em}{\includegraphics[width=\linewidth]{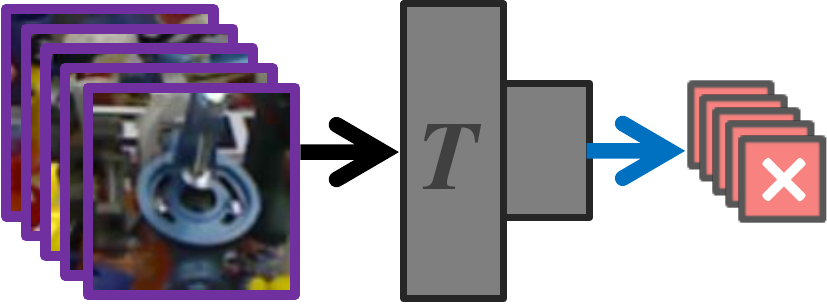}
}
\\[2pt]
\cline{2-4}  \rule[5pt]{0pt}{23pt} 

& \textit{PixelDA}~\cite{bousmalis2016unsupervised} &
\parbox[c]{10em}{\includegraphics[width=\linewidth]{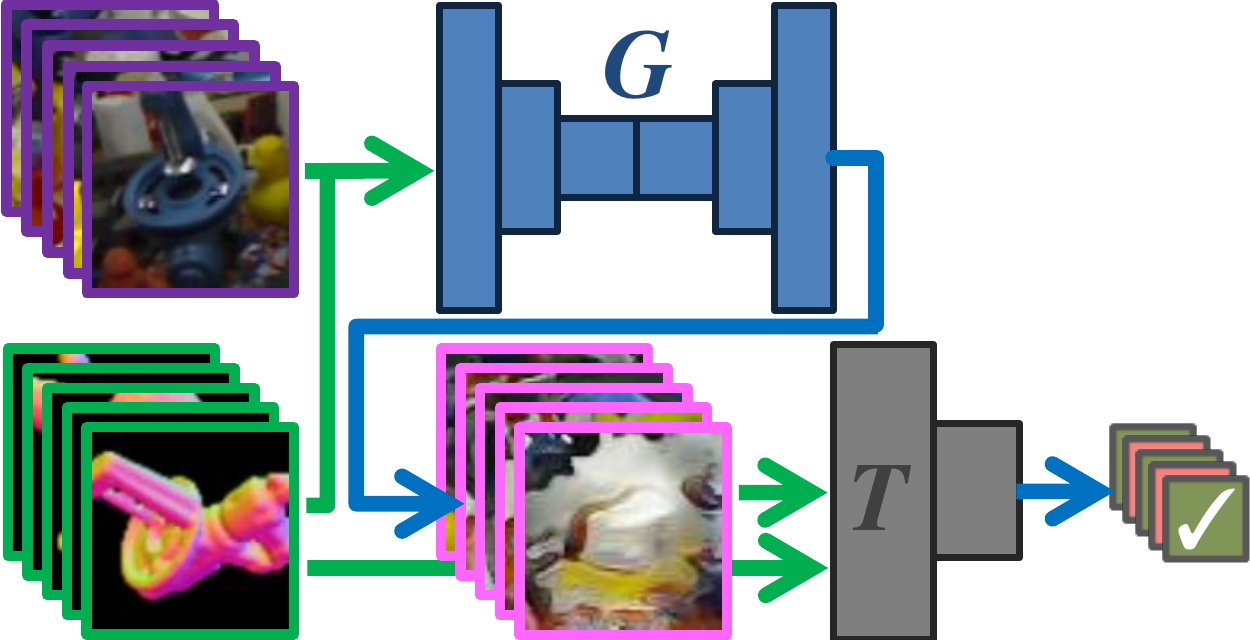}
}
&
\parbox[c]{6.61417em}{\includegraphics[width=\linewidth]{methods/real_rgb+notex_test_pixelda}
}
\\
\midrule

\multirow{2}{*}{\parbox[c]{8em}{
\includegraphics[width=\linewidth]{symbol_cad+tex2}}
}	
& supervised &
\parbox[c]{6.61417em}{\includegraphics[width=\linewidth]{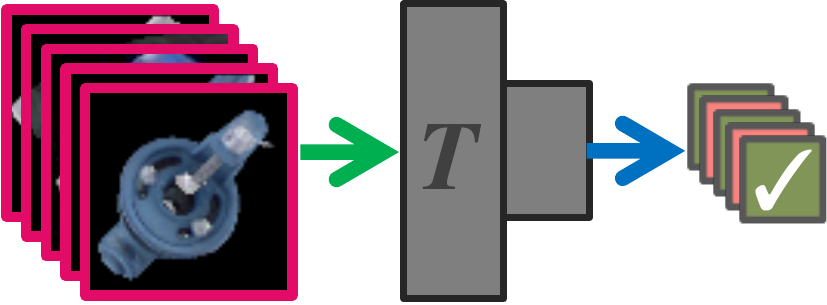}
}
&
\parbox[c]{6.61417em}{\includegraphics[width=\linewidth]{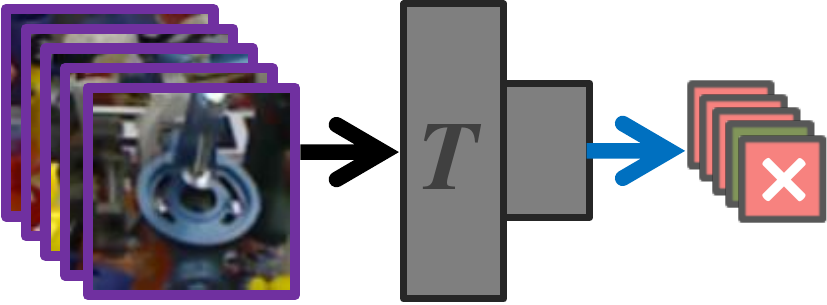}
} 
\\[2pt]
\cline{2-4}  \rule[5pt]{0pt}{12pt} 

& \parbox[c]{4.8em}{\shortstack{supervised\\$+$ augment.}} &
\parbox[c]{10em}{\includegraphics[width=\linewidth]{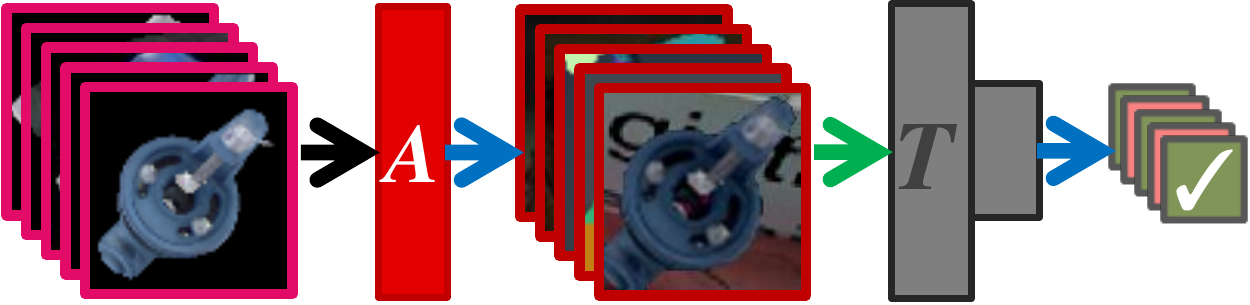}
}
&
\parbox[c]{6.61417em}{\includegraphics[width=\linewidth]{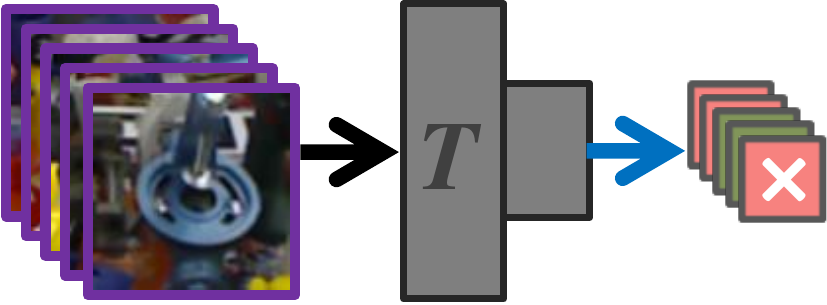}
}
\\
\midrule

\parbox[c]{8em}{
\includegraphics[width=\linewidth]{symbol_real2+cad2}
}	
&  \parbox[c]{4.4em}{\shortstack{supervised\\$+$ generic\\ regressor\\~\cite{laina2016deeper}}} &
\parbox[c]{8.503937em}{\includegraphics[width=\linewidth]{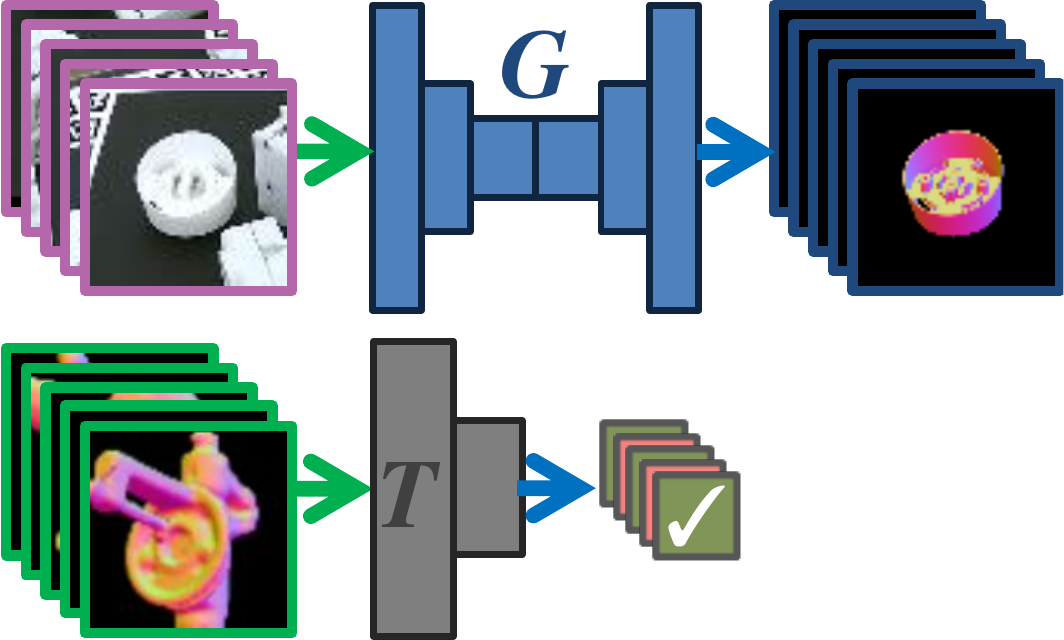}
}
&
\parbox[c]{6.61417em}{\includegraphics[width=\linewidth]{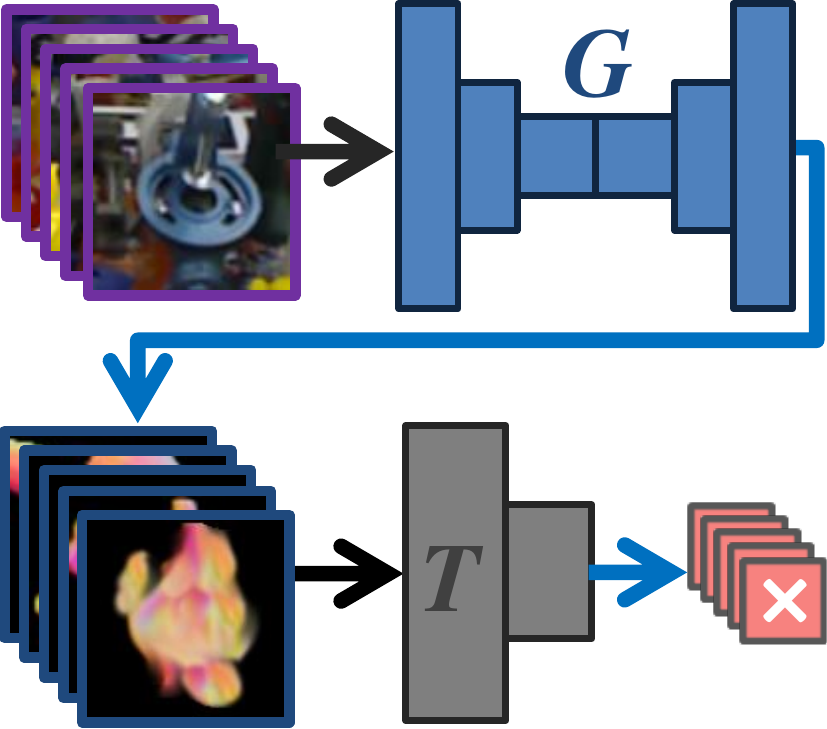}
}
\\
\midrule

\multirow{2}{*}{\parbox[c]{8em}{
\includegraphics[width=\linewidth]{symbol_cad2}}
}	
&  \parbox[c]{4.8em}{\shortstack{supervised\\$+$ augment.}} &
\parbox[c]{10em}{\includegraphics[width=\linewidth]{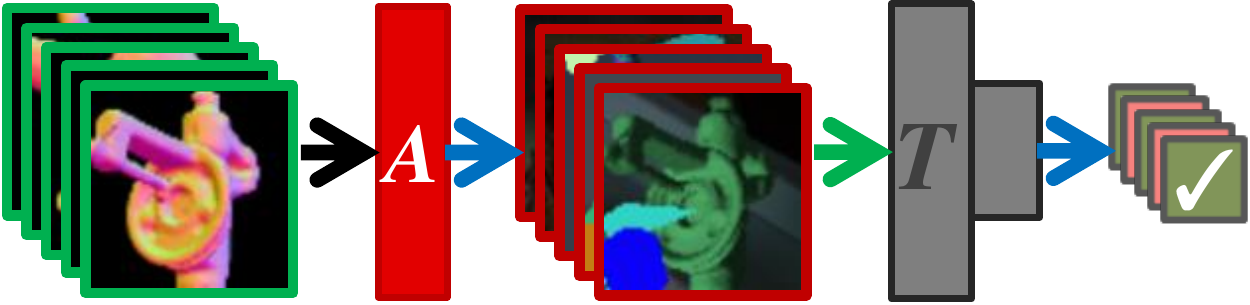}
}
&
\parbox[c]{6.61417em}{\includegraphics[width=\linewidth]{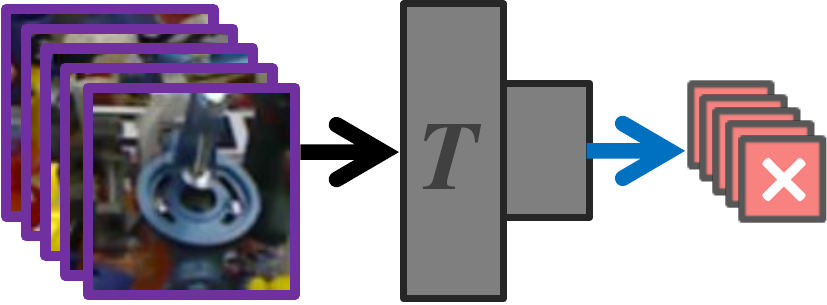}
}
\\[2pt]
\cline{2-4}  \rule[15pt]{0pt}{18pt} 

& \shortstack{\textit{SynDA}\\(ours)} &
\parbox[c]{10em}{\includegraphics[width=\linewidth]{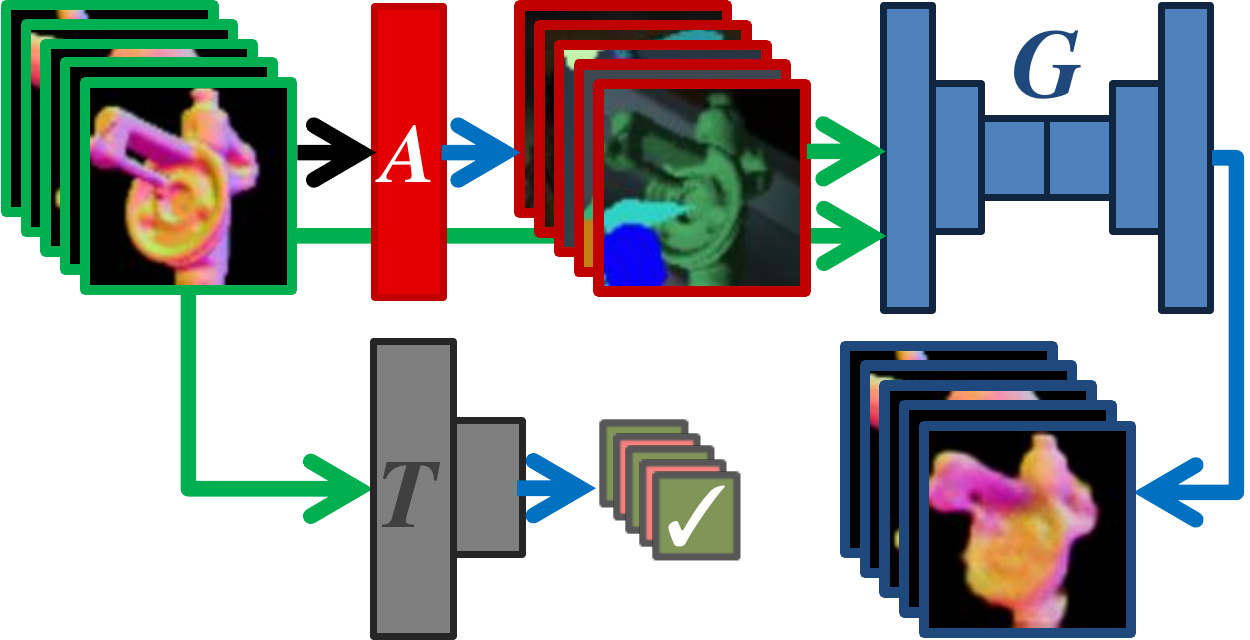}
}
&
\parbox[c]{6.61417em}{\includegraphics[width=\linewidth]{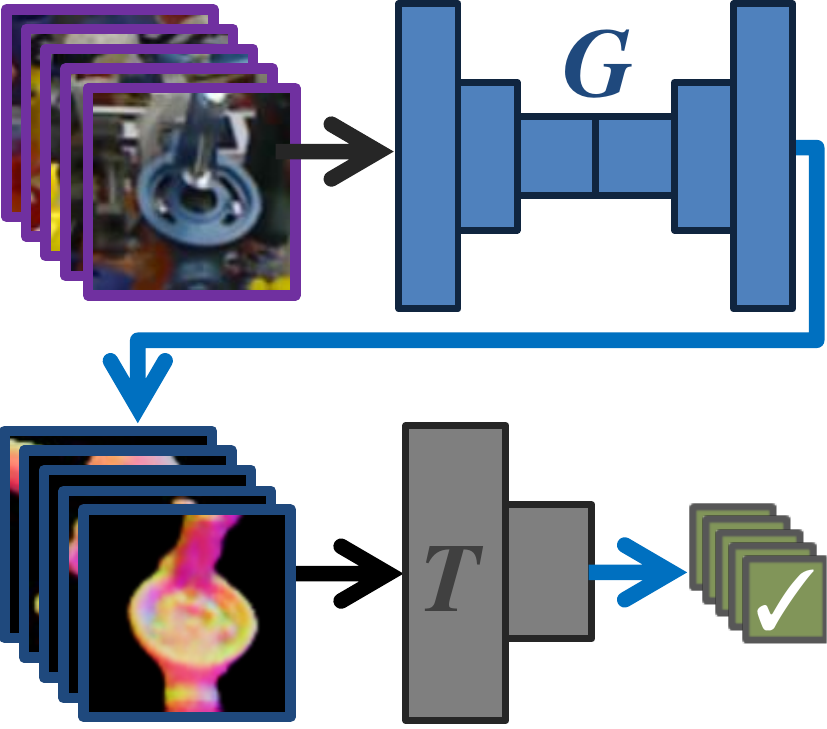}
}

\\
\bottomrule
\\[-1em]
\multicolumn{4}{c}{
\parbox[c]{34em}{
\includegraphics[width=\linewidth]{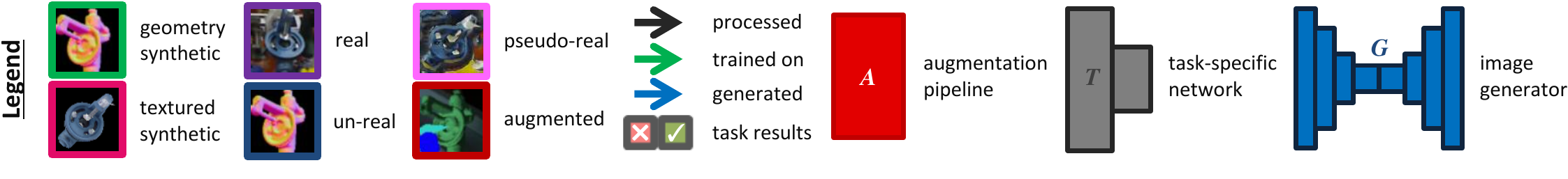}
}
}
\end{tabu}
}
\end{table*}

\section{Schematic Overview of the Different Gap-Bridging Methods}

Table~\ref{tab:comp} contains a schematic comparison of the training and testing solutions for recognition tasks addressed in the paper, depending on the type of data available for training (real images, corresponding annotations, CAD models, corresponding realistic textures, or real images from a different domain). 

While usual domain adaptation methods such as \textit{PixelDA}~\cite{bousmalis2016unsupervised} or \textit{DANN}~\cite{ganin2016domain} focus on use-cases when real-world data are available---in terms or unlabeled target images but also realistic textures for 3D CAD models---we develop our solution on the assumption of minimal information on the target domain. For CAD-based recognition applications, we demonstrate in the main paper how \textit{SynDA} yields state-of-the-art results when only pure geometrical data is available. For comparison, both \textit{PixelDA}~\cite{bousmalis2016unsupervised} and \textit{DANN}~\cite{ganin2016domain} fail to map real and synthetic domains when provided with texture-less rendered images \textit{and} real pictures, as illustrated in Figure~\ref{fig:pixelda}.

\section{Implementation Details}

This section contains more in depth details on network architecture and parameters, augmentation pipeline, as well as synthetic data generation.

	\subsection{Network Architecture and Parameters}
	
Figure~\refwithdefault{fig:architecture}{3} of the paper already provides the readers with an exhaustive overview of our state-of-the-art architecture, layer by layer.
Our solution is implemented in Python using the TensorFlow framework~\cite{abadi2016tensorflow}.
	
\paragraph{Layer parameterization:}
	\begin{itemize}
		\setlength\itemsep{0em}
		\item All Convolution layers have $4 \times 4$ filter kernels;
		\item All Dropout layers have a dropout rate of $50\%$;
		\item All LeakyReLU layers have a leakiness of $0.2$;
		\item Input and output images are $64 \times 64$px, normalized between $-1$ and $1$.
	\end{itemize}
	
\paragraph{Training parameters:}
	\begin{itemize}
		\setlength\itemsep{0em}
		\item Weights are initialized from a zero-centered Gaussian distribution, with a standard deviation of $0.02$ ;
		\item The Adam optimizer~\cite{kingma2014adam} is used, with $\beta_1 = 0.5$;
		\item The base learning rate is initialized at $2e^{-4}$.
	\end{itemize}

	\subsection{Augmentation Pipeline Details}

\par\noindent
\textbf{Simple random shading:}
To generate a virtually infinite dataset of color images, we transform our set of geometrical maps (\ie normal maps) into random images with an online augmentation procedure. Its main and first operation thus consists in shading, \ie generating lightness maps from the normal maps, sampling random light conditions. It is done through a custom Blinn-Phong shading~\cite{blinn1977models} method, as described in Algorithm~\ref{alg:shading}.

Lighting parameters, defined through the augmentation noise vector $z$, are sampled using uniform distributions \eg 
$\mathcal{U}(0.05, 0.3)$ for each color components of $a$, $\mathcal{U}(0.1, 0.8)$ for each component of $d$, $\mathcal{U}(0, 0.1)$ for each component of $s$, $\mathcal{U}(0.9, 1.1)$ for each component of $s_p$, \etc.

	\begin{algorithm}[t]
		\KwIn{ 
			$N \in \mathbb{R}^{h \times w \times 3}$ normal map,
			$L \in \mathbb{R}^3$ directional light vector,
			$a \in \mathbb{R^3}$ RGB ambient light coefficient,
			$d \in \mathbb{R^3}$ RGB diffusion coefficient,
			$s \in \mathbb{R}^3$ RGB specularity coefficient,
			$s_p \in \mathbb{R}$ specular hardness,
			$f_x \in \mathbb{R}^2$ pixel focal range used to render $N$

		}
		\KwOut{ 
			$M \in \mathbb{R}^{h \times w \times 3}$ color lightness map
				
		}
		
		\tcc{- Simplification \#1: we recover an approximate viewer vector $V$ from $N$ indices and $f_x$.}
		\tcc{- Simplification \#2: we suppose the light source at $+\infty$ distance, hence the same $L$ for every surface point.}
		\tcc{- Note: we use Einstein notation for matrix-vector operations.}
		\tcc{}
		\tcc{View vector approximation:}
		$V \leftarrow \big\{ (j, i, 1) \big\}_{j=0,i=0}^{h,w}$\;
		$V_j \leftarrow -\frac{(V_j - \sfrac{h}{2})}{f_{x,j}} ; V_i \leftarrow -\frac{(V_i - \sfrac{w}{2})}{f_{x,i}}$\;
		$V \leftarrow \frac{V}{\|V\|}$\;
		
		\tcc{Computation of half-way vector map:}
		$H \leftarrow V + L$;
		
		\tcc{Diffuse shading:}
		$D^{ij} \leftarrow N^{ij}{}_k L^k$\;
		
		\tcc{Specular shading:}
		$S^{ij} \leftarrow (N^{ij}{}_k H^k{}_{ij})^{s_p}$\;
		
		\tcc{Adding all contributions (given $e_{ijc} = e_i \otimes e_j \otimes e_c$) :}
		$M^{ijc} \leftarrow \min\big(\max(a \cdot e_{ijc} + d \cdot D^{ij}e_c + s \cdot S^{ij}e_c  \ , 0)  \ , 1\big)$\;
		
		\KwRet{$M$}
		
		\caption{Approximate Blinn-Phong shading~\cite{blinn1977models} from normal maps}
		\label{alg:shading}
	\end{algorithm}

\par\noindent
\textbf{Stochastic texturing:}
Random textures are applied to the objects every iteration. 
They are noise-generated, using the open-source FastNoise library~\cite{fastnoise}. In particular, Perlin noise~\cite{perlin2002improving}, cellular noise~\cite{worley1996cellular}, and white noise are used, sampled from the uniform distribution $\mathcal{U}(0.0001,0.1)$, to obtain hue and saturation maps.
These maps are either directly merged to the lightness map of the objects to obtain their final HSL appearance, or either used as texture maps. In the second case, the original surface normals are projected on two dimensions (randomly dropping their $X$, $Y$ or $Z$ axis), to be used as an UV map for the texture.

\par\noindent
\textbf{Background addition:}
As mentioned in the paper, random backgrounds are added, either generated using the aforementioned noise sources or obtained by randomly cropping and resizing color images from public datasets (\eg COCO~\cite{lin2014microsoft}). 
The background pixel values are multiplied by the normalized foreground brightness, to get more homogenized results.

	\par\noindent
	\textbf{Random occlusions:}
	To generate occlusions we use a function from \cite{ounsworth2015anticipatory}, where they generate 2D obstacles for a drone moving planning simulation. The points $p$ are sampled by walking around the circle taking random angular steps and random radii at each step. A polygon is then generated using the points $p$ and filled with either random noise (\cf main paper) or textures if provided.

The polygon's complexity is defined by two parameters:  $\sigma$ ("spikeyness"), which controls how much point coordinates vary from the radius $r_{ave}$, and $\epsilon$ ("irregularity"), which sets an error to the default uniform angular distribution. 
Variables $c_X$ and $c_Y$ define the polygon center; $r_{ave}$ its average radius; $\delta \theta$ and $\theta$ are a vector of angle steps, and a vector of angles respectively; and $l_x \times l_y$ are the image dimensions (equal to $64  \times 64$ px here).   The pseudocode is listed in Algorithm~\ref{alg:occlusion}. 

\begin{algorithm}[t]
	\KwIn{ 
			$z \in \mathbb{Z}^k$ noise vector
		}
		\KwOut{ 
			$p = \{p_i \in \mathbb{R}^2\}_{i=0}^{N_{vert}}$ polygon points	
		}
		
		\tcc{occlusion parameters sampling:}
		$c_{x}, c_{y}, r_{ave}, N_{vert}, \epsilon , \sigma  \leftarrow \text{sampleFromVector2}(z)$\;

		\tcc{angle steps generation:}
		$sum = 0$ \;
		\For{$ i \in \{1, \ldots, N_{vert}\} $}{
			$\delta\theta_{i} \leftarrow \mathcal{U}(\sfrac{2 \pi}{N_{vert}} - \epsilon, \sfrac{2 \pi}{N_{vert}} + \epsilon)$ \;
			$sum \leftarrow sum + step$ \;
		}
		\tcc{steps normalization:}
		$k \leftarrow \sfrac{sum}{(2 \pi)}$ \;
		\For{$ i \in \{1, \ldots, N_{vert}\}$}{
			$\delta\theta_i \leftarrow \sfrac{\delta\theta_i}{k}$ \;
		}
		\tcc{polygon points generation:}
		$\theta_1 \leftarrow \mathcal{U}(0, 2 \pi)$ \;
		\For{$ i \in \{1, \ldots, N_{vert}\} $}{
			$r \leftarrow \mathcal{N}(r_{ave}, \sigma)$ \;
			$p_i \leftarrow (c_X + r \cos(\theta_i), c_Y + r \sin(\theta_i))$ \;
			$\theta_i \leftarrow \theta_i + \delta\theta_i$
		}
		\KwRet{p}
		
		\caption{Random polygon generation \cite{ounsworth2015anticipatory}}
		\label{alg:occlusion}
	\end{algorithm}
	
	Occlusion parameters are also set by the noise vector $z$. In our experiments, the following sampling distributions are used (with $\mathcal{B}$ -- Bernoulli, $\mathcal{U}$ -- Uniform, and $\mathcal{N}$ -- Gaussian): 
\begin{itemize}
\setlength\itemsep{0em}
\item $\mathcal{B}\big(\mathcal{U}(0, \sfrac{l_x}{4}),\,\mathcal{U}(\sfrac{l_x}{4}, l)\big)$ for $c_X$, $\mathcal{B}\big(\mathcal{U}(0, \sfrac{l_y}{4}),\,\mathcal{U}(\sfrac{l_y}{4}, l)\big)$ for $c_Y$;
\item $\mathcal{U}(10, \sfrac{l}{4})$ for $r_{ave}$, with $l = \min(l_x, l_y)$;
\item $\mathcal{U}(3, 10)$ for $N_{vert}$;
\item $\mathcal{U}(0, 0.5)$ for $\sigma$.
\end{itemize}

	\subsection{Synthetic Data Generation}
	
	Subsection 3.2 of the main paper gives an overview of the data generation method used for our pipeline. In order to produce normal maps of objects of interest for our augmentation pipeline, we use OpenGL~\cite{opengl} with a custom normal shader printing its output in 3 output channels. Viewpoint are defined by the verteces of an icosahedron centered on target objects. In order to achieve a finer sampling, one needs to subsequently subdivide triangular faces of the icosahedron to smaller triangles until the desired level of detail is achieved. In-plane rotations can also be introduced by rotating the camera around the ray pointing to the object. 
	
	For the T-LESS dataset~\cite{hodan2017t}, synthetic data is rendered using the full icosahedron with a radius of 600mm and 3 subdivisions since real sequence contain objects shown from below as well. No in-plane rotations were added to the training data neither due to the same reason.
	This resulted in generation of 642 samples per object (given 11 objects --- numbers 2, 5, 6, 7, 8, 11, 12, 18, 25, 29 and 30).

\begin{figure}[t]
		\centering
		
		\subfigure[Axis symmetric]
			{\includegraphics[width=.32\linewidth]{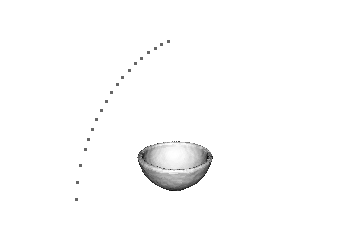}\label{fig:axis}}
		 \hfill
		\subfigure[Plane symmetric]
			{\includegraphics[width=.32\linewidth]{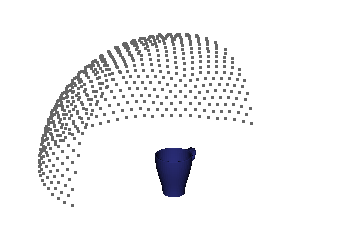}\label{fig:plans}}
		 \hfill
		\subfigure[Regular]
			{\includegraphics[width=.32\linewidth]{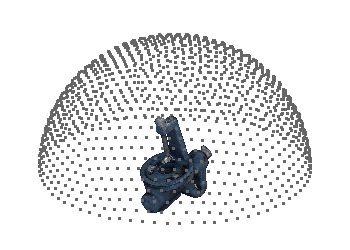}\label{fig:regular}}

		\caption{\textbf{Vertices sampling for different LineMOD objects} -- each vertex represents a camera position from which the object is rendered.}
		\label{fig:samplings}
	\end{figure}

	LineMOD data was generated using an icosahedron of radius 600mm with 3 consecutive subdivisions. Only the upper part of the icosahedron is used since in real sequences all objects are shot from above. In-plane rotations also added for each vertex, parametrized from -$45^\circ$ to $45^\circ$ with a stride of $15^\circ$. 
	Rotation invariance of four irregular LineMOD objects (\textit{bowl}, \textit{cup}, \textit{eggbox}, and \textit{glue}) was also taken into account by limiting the amount of sampling points, such that each patch is unique. Figure~\ref{fig:samplings}  demonstrates the results of the output vertex sampling for different object types. We therefore generated 2,359 data points for each of the 11 regular objects, 1,239 for 3 plane symmetric objects (\textit{cup}, \textit{glue}, and \textit{eggbox}) and 119 for the axis symmetric \textit{bowl\textit{}}.

\section{Additional Qualitative Results}

\begin{figure*}[h]
\centering
\includegraphics[width=1\linewidth]{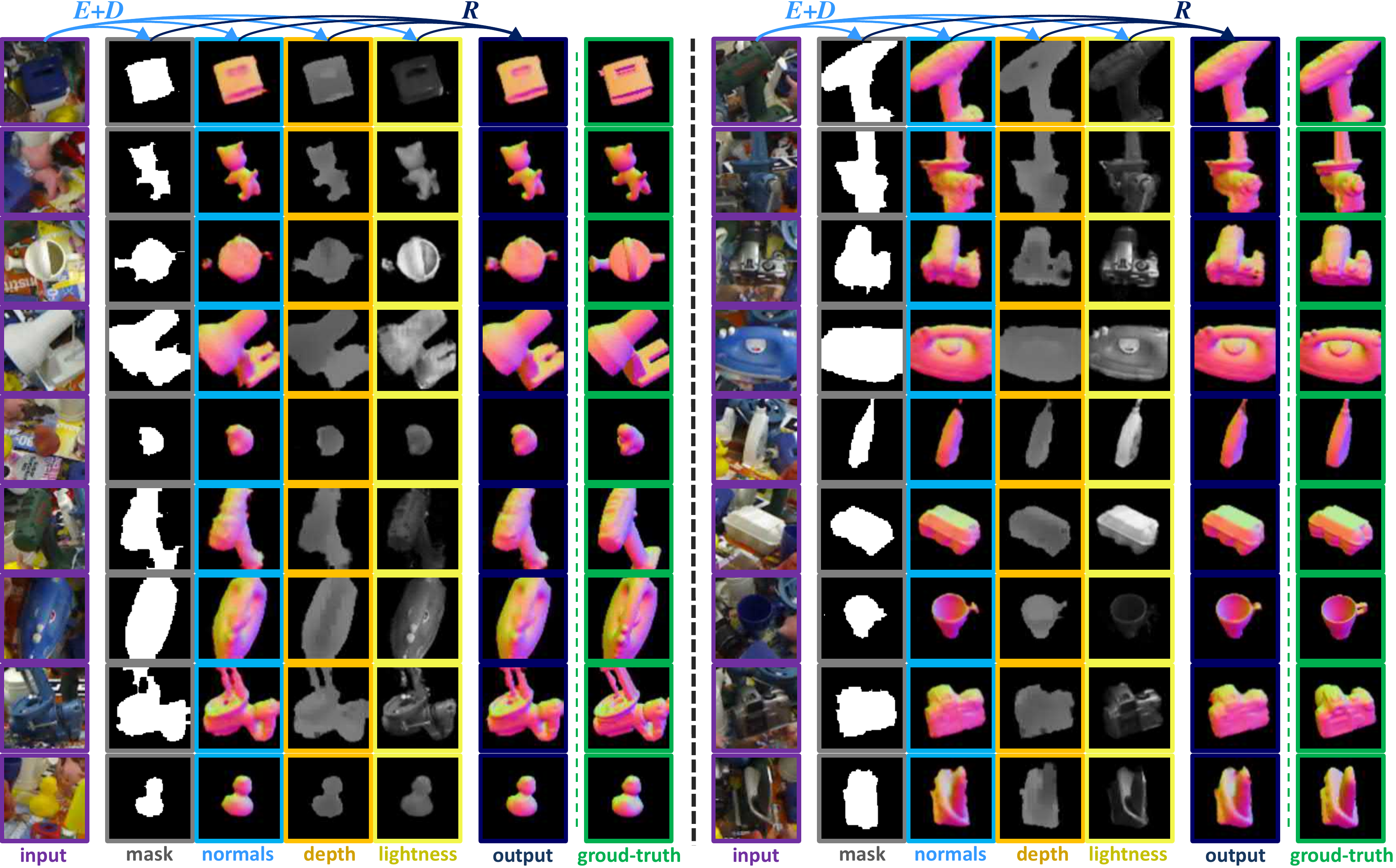}
\caption{\textbf{Qualitative results on LineMOD}~\cite{hinterstoisser2012model} for \textit{SynDA} trained on texture-less CAD data.}
\label{fig:sup_linemod}  
\end{figure*}

\begin{figure*}[h]
\centering
\includegraphics[width=1\linewidth]{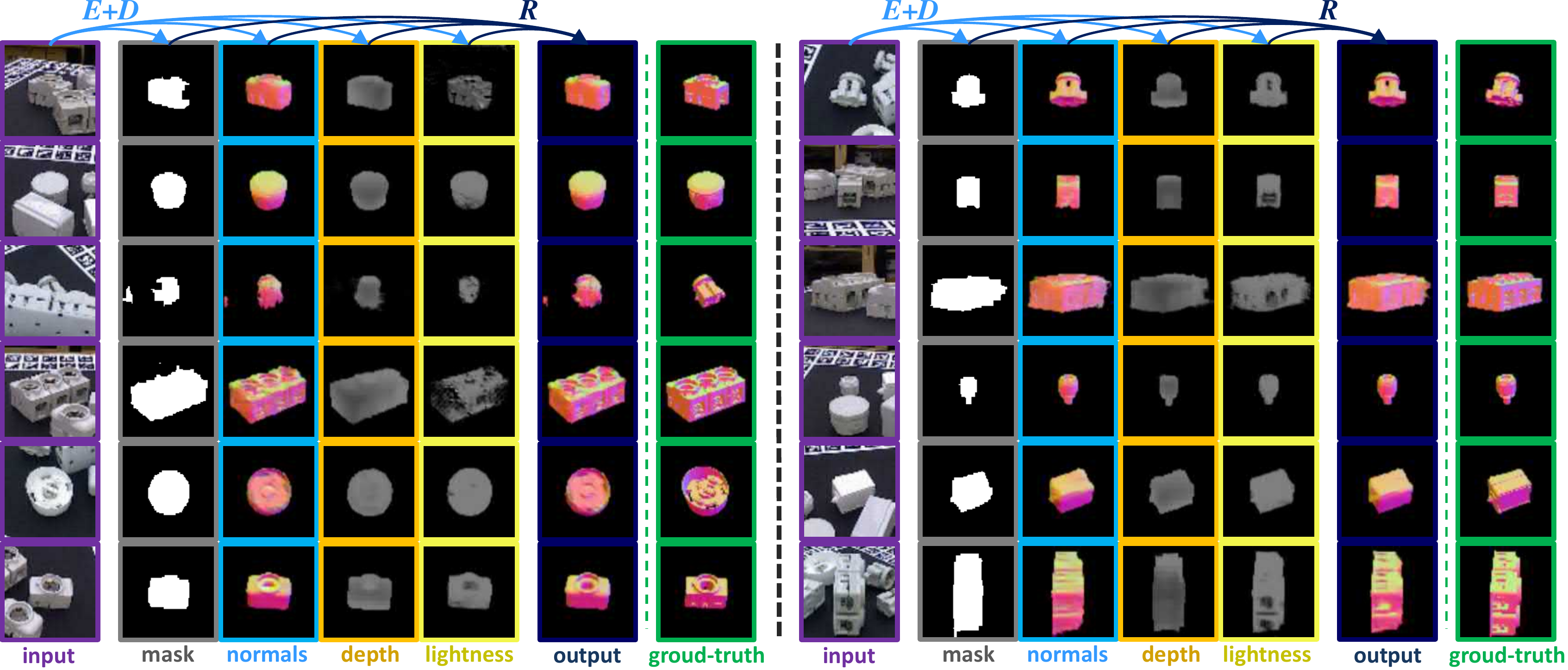}
\caption{\textbf{Qualitative results on T-LESS}~\cite{hodan2017t}, for \textit{SynDA} trained on texture-less CAD data.}
\label{fig:sup_tless}  
\end{figure*}

Figures~\ref{fig:sup_linemod} and~\ref{fig:sup_tless} contain further visual results, demonstrating how our pipeline fairs on real color images when trained purely on synthetic geometrical data.

\clearpage
\clearpage

{\small
\bibliographystyle{ieee}
\bibliography{references}
}

\end{document}